\documentclass[conference]{IEEEtran}
\IEEEoverridecommandlockouts

\usepackage{cite}
\usepackage{amsmath,amssymb,amsfonts}
\usepackage{algorithmic}
\usepackage{graphicx}
\usepackage{textcomp}
\usepackage{xcolor}
\def\BibTeX{{\rm B\kern-.05em{\sc i\kern-.025em b}\kern-.08em
    T\kern-.1667em\lower.7ex\hbox{E}\kern-.125emX}}

\usepackage{amsmath,amsfonts,bm}

\def\eqref#1{equation~\ref{#1}}

\def\1{\bm{1}}

\def\vmu{{\bm{\mu}}}
\def\vtheta{{\bm{\theta}}}

\def\vb{{\bm{b}}}

\def\vx{{\bm{x}}}
\def\vy{{\bm{y}}}
\def\vz{{\bm{z}}}

\def\mA{{\bm{A}}}

\def\mI{{\bm{I}}}

\def\mW{{\bm{W}}}
\def\mX{{\bm{X}}}

\def\mSigma{{\bm{\Sigma}}}

\DeclareMathAlphabet{\mathsfit}{\encodingdefault}{\sfdefault}{m}{sl}
\SetMathAlphabet{\mathsfit}{bold}{\encodingdefault}{\sfdefault}{bx}{n}
\newcommand{\tens}[1]{\bm{\mathsfit{#1}}}
\def\tA{{\tens{A}}}

\def\tT{{\tens{T}}}

\newcommand{\etens}[1]{\mathsfit{#1}}

\def\etA{{\etens{A}}}
\def\etB{{\etens{B}}}

\newcommand{\E}{\mathbb{E}}

\newcommand{\R}{\mathbb{R}}

\usepackage{amsmath}
\usepackage{amssymb}
\usepackage{hyperref}
\usepackage{cleveref}
\usepackage{bm}
\usepackage[acronym,nomain,shortcuts]{glossaries}
\usepackage{url}
\usepackage{layouts}
\usepackage{enumitem}
\usepackage{diagbox}
\usepackage[export]{adjustbox}
\usepackage{siunitx}
\usepackage{comment}
\usepackage{multirow}
\usepackage{hhline}
\usepackage{booktabs}
\usepackage{makecell}
\usepackage{float}
\usepackage{subcaption}
\glsdisablehyper
\usepackage{stfloats} 

\newcommand{\enquote}[1]{#1}

\newcommand{\vbeta}{{\bm{\beta}}}
\newcommand{\vpi}{{\bm{\pi}}}
\newcommand{\cen}[1]{\multicolumn{1}{c|}{#1}}
\newacronym{GMM}{GMM}{Gaussian Mixture Model}
\newacronym{DCGMM}{DCGMM}{Deep Convolutional Gaussian Mixture Model}
\newacronym{CNN}{CNN}{Convolutional Neural Network}
\newacronym{GAN}{GAN}{Generative Adverserial Network}
\newacronym{CGAN}{CGAN}{Convolutional Generative Adverserial Network}
\newacronym{SGD}{SGD}{Stochastic Gradient Descent}
\newacronym{DNN}{DNN}{Deep Neural Network}
\newacronym{VAE}{VAE}{Variational AutoEncoder}
\newacronym{MFA}{MFA}{Mixture of Factor Analyzer}
\newacronym{EM}{EM}{Expectation-Maximization}
\newacronym{SLT}{SLT}{Sequential Learning Task}
\newacronym{FC}{FC}{Fitting Capacity}
\newacronym{GPU}{GPU}{Graphics Processing Unit}
\newacronym{RF}{RF}{Receptive Field}
\newacronym{F}{F}{Folding}
\newacronym{G}{G}{GMM}
\newacronym{P}{P}{Pooling}
\newacronym{C}{C}{Linear Classifier}

\let\oldsqrt\sqrt
\def\sqrt{\mathpalette\DHLhksqrt}
\def\DHLhksqrt#1#2{%
	\setbox0=\hbox{$#1\oldsqrt{#2\,}$}\dimen0=\ht0
	\advance\dimen0-0.2\ht0
	\setbox2=\hbox{\vrule height\ht0 depth -\dimen0}%
	{\box0\lower0.4pt\box2}}

\crefname{section}{Sec.}{Secs.}
\crefname{subsection}{Sec.}{Secs.}
\crefname{subsubsection}{Sec.}{Secs.}
\crefname{paragraph}{Par.}{Para.}
\crefname{appendix}{App.}{Apps.}
\crefname{table}{Tab.}{Tabs.}
\crefname{figure}{Fig.}{Figs.}
\crefname{equation}{Eq.}{Eq.}
\Crefname{equation}{Eq.}{Eq.}
\crefname{algorithm}{Alg.}{Algs.}

\makeatletter
\def\ps@IEEEtitlepagestyle{%
	\def\@oddfoot{\mycopyrightnotice}%
	\def\@oddhead{\hbox{}\@IEEEheaderstyle\leftmark\hfil\thepage}\relax
	\def\@evenhead{\@IEEEheaderstyle\thepage\hfil\leftmark\hbox{}}\relax
	\def\@evenfoot{}%
}
\def\mycopyrightnotice{%
	\begin{minipage}{\textwidth}
		\centering \scriptsize
		Copyright~\copyright~2021 IEEE. Personal use of this material is permitted. Permission from IEEE must be obtained for all other uses, in any current or future media, including\\reprinting/republishing this material for advertising or promotional purposes, creating new collective works, for resale or redistribution to servers or lists, or reuse of any copyrighted component of this work in other works by sending a request to pubs-permissions@ieee.org.
	\end{minipage}
}
\makeatother

\begin{document}

\title{Image Modeling with\\Deep Convolutional Gaussian Mixture Models}

\author{\IEEEauthorblockN{Alexander Gepperth}
	\IEEEauthorblockA{\textit{Fulda University of Applied Sciences}\\
		Fulda, Germany \\
		alexander.gepperth@cs.hs-fulda.de}
	\and
	\IEEEauthorblockN{Benedikt Pf{\"u}lb}
	\IEEEauthorblockA{\textit{Fulda University of Applied Sciences}\\
		Fulda, Germany \\
		benedikt.pfuelb@cs.hs-fulda.de}
}

\maketitle

\begin{abstract}
In this conceptual work, we present \acp{DCGMM}: a new formulation of deep hierarchical \acp{GMM} that is particularly suitable for describing and generating images.
Vanilla (i.e., \enquote{flat}) \acp{GMM} require a very large number of components to describe images well, leading to long training times and memory issues. 
\Acp{DCGMM} avoid this by a stacked architecture of multiple \ac{GMM} layers, linked by convolution and pooling operations. 
This allows to exploit the compositionality of images in a similar way as deep \acsp{CNN} do.
\Acp{DCGMM} can be trained end-to-end by \acl{SGD}.
This sets them apart from vanilla \acp{GMM} which are trained by \acl{EM}, requiring a prior k-means initialization which is infeasible in a layered structure.
For generating sharp images with \acp{DCGMM}, we introduce a new gradient-based technique for sampling through non-invertible operations like convolution and pooling.
Based on the MNIST and FashionMNIST datasets, we validate the \acp{DCGMM} model by demonstrating its superiority over \enquote{flat} \acp{GMM} for clustering, sampling and outlier detection.
\end{abstract}
\vspace{.5em}
\section{Introduction}
\noindent This conceptual work is in the context of probabilistic image modeling, whose main objectives are both, density estimation and image generation (sampling). 
Since images usually do not precisely follow a Gaussian mixture distribution, such a treatment is inherently approximative in nature.
Image generation is currently a very active research topic, and similar techniques are being investigated for generating videos \cite{10.1007/978-3-030-00934-2_31,pmlr-v97-piergiovanni19a}. 
\par
An issue with many recent approaches is the lack of density estimation capacity, i.e., explicitly expressing the learned probability-under-the-model $p(\vx)$ of an image $\vx$. 
\par 
In contrast, \acp{GMM} explicitly describe the distribution $p(\mX)$, given by a set of training data $\mX$\,$=$\,$\{\vx_n\}$, as a weighted mixture of $K$ Gaussian component densities $\mathcal{N}(\vx;\vmu_k,\mSigma_k)$\,$\equiv$\,$\mathcal{N}_k(\vx)$: 
\begin{align}
	\label{eqn:prob:incomplete}
	p(\vx) & =\sum_k^K \pi_k \mathcal{N}_k(\vx).
\end{align}
Conceptually, GMMs are latent-variable models: it is assumed that the unobservable (latent) variable $z$ determines from which component a data vector $\vx$ has been sampled, which is expressed as
\begin{align}
	\label{eqn:prob:complete}
	p(\vx,z) = \pi_z \mathcal N_z(\vx)
\end{align}
Marginalizing the latent variable in \cref{eqn:prob:complete}, we obtain \cref{eqn:prob:incomplete}.

\Acp{GMM} require the mixture weights to be normalized: $\sum_k \pi_k$\,$=$\,$1$ and the covariance matrices to be positive definite: $\vx^T \mSigma_k \vx$\,$>$\,$0$\ $\forall \vx$.
The quality of the current \enquote{fit-to-data} is expressed by the (incomplete) log-likelihood
\begin{equation}
	\mathcal{L} (\mX)= \E_n\left[ \log \sum_k \pi_k \mathcal{N}_k(\vx_n)\right],
	\label{eqn:loglikelihood}
\end{equation}
which is what \ac{GMM} training optimizes, usually by variants of \ac{EM} \cite{Dempster1977}.
It is shown that any distributions can be approximated by mixtures of Gaussians given enough components \cite{Goodfellow2016}.
Thus, \acp{GMM} are guaranteed to model the complete data distribution, but only to the extent allowed by the number of components~$K$.
\par 
In this respect, \acp{GMM} are similar to \enquote{flat} neural networks with a single hidden layer.
Although, by the universal approximation theorem of \cite{Pinkus1999} and \cite{Hornik1989}, they \textit{can} approximate arbitrary functions (from certain rather broad function classes), they fail to do so in practice.
The reason for this is that the number of required hidden layer elements is unknown, and usually beyond the reach of any reasonable computational capacity.
For images, this problem was largely solved by introducing deep \acp{CNN}.
\Acp{CNN} model the statistical structure of images (hierarchical organization and translation invariance) by chaining multiple convolution and pooling layers.
Thus, the number of parameters can be reduced without compromising accuracy.
\subsection{Objective, Contribution and Novelty}
\noindent The objective of this article is to introduce a \ac{GMM} architecture which exploits the same principles that led to the performance explosion of \acp{CNN}. 
In particular, the genuinely novel characteristics are:
\begin{itemize}[leftmargin=*]
	\setlength\itemsep{0em}
	\item conceptually new formulation of deep GMMs, including convolution and pooling operations
	\item scalable end-to-end training by SGD from random initial conditions (no k-means initialization)
  \item efficient large-scale training on high-demensional images
	\item realistic sampling despite non-invertible operations (pooling, convolutions)
	\item better empirical performance than vanilla \acp{GMM} on images for sampling, clustering and outlier detection
\end{itemize}
In addition, we provide a publicly TensorFlow implementation which supports a Keras-like construction of DCGMMs.
\subsection{Related Work}\label{sec:relwork}
\noindent\textbf{GANs and VAEs}\hspace{.5em}
The currently most widely used models of image modeling and generation are \acp{GAN} \cite{Arjovsky2017,Mirza2014,Goodfellow2014} and \acp{VAE}. 
\Acp{GAN} are capable of sampling photo-realistic images \cite{Richardson2018}, but are unable to perform density estimation since there is no way to obtain $p(\vx)$ for a given sample $\vx$. 
Furthermore, their probabilistic interpretation remains unclear since they do not possess a differentiable loss function that is minimized by training. 
They may suffer from what is termed \textit{mode collapse}, which is hard to detect automatically due to the absence of a loss function \cite{Richardson2018}.
\Acp{VAE} show similar performance when it comes to sampling, in addition to minimizing a differentiable loss function. 
Moreover, density estimation with \acp{VAE} is problematic. 
Similar approaches, with similar strengths in sampling but a lack of density estimation, are realized by the PixelCNN architecture\cite{oord2016conditional}, GLOW\cite{kingma2018glow} and their variants.
\smallskip
\par\noindent\textbf{Hierarchical \acp{GMM}}\hspace{.5em}
A hierarchical extension of \acp{GMM} is presented by \cite{Liu2002} with the goal of unsupervised clustering: responsibilities of one \ac{GMM} are treated as inputs to a subsequent \ac{GMM}, together with an adaptive mechanism that determines the depth of the hierarchy.
\cite{Garcia2010} present a comparable, more information-theoretic approach but not targetting sampling either.
The closest related work to the model we present here is described in \cite{Viroli2019,VanDenOord2014,Tang2012}. 
All of these models perceive hierarchical GMMs as modular decompositions of GMMs that are flat. 
The conceptual foundation is that sampling from a GMM implies the transformation of a normally distributed latent variable $\vz$ by a single GMM component $k$, with weight $\pi_k$,  $\vz$ as $\vy$\,$=$\,$\mA_k \vz$\,$+$\,$\vbeta_k$\,$+$\,$\epsilon_k$. 
This leads to a distribution $\vy$\,$\sim$\,$\mathcal N(\vbeta_k,\mA_k\mA_k^\top$\,$+$\,$\epsilon_k)$. 
The choice which GMM component is \enquote{allowed} to transform the latent variable is based on component weights $\pi_k$. 
Hierarchical GMMs are now realized by \textit{several} sampling steps (layers), and the sampling results of the previous layer representing the latent variable to be transformed by the next one.
None of these models consider convolutional or max-pooling operations which have been proven to be important for modeling the statistical structure of images.
\smallskip\par
\noindent\textbf{\acp{MFA}}\hspace{.5em}
\Acp{MFA} models \cite{McLachlan2005,Ghahramani1997} can be considered as hierarchical \acp{GMM} because they are formulated in terms of a lower-dimensional latent-variable representation, which is mapped to a higher-dimensional space.
The use of \acp{MFA} for describing natural images is discussed in detail in \cite{Richardson2018}, showing that the \ac{MFA} model alone, without further hierarchical structure, compares quite favorably to \acp{GAN} when considering image generation. 
\smallskip\par
\noindent\textbf{Hierarchical Mixture Models}\hspace{.5em}
An interesting overview of the current research landscape in terms of hierarchical generative mixture models is given in \cite{jaini2018deep}. 
All of these models are, in principle, capable of sampling and density estimation although the quality of sampling, and the data they can be trained on, vary considerably.
Principal competitors in this domain are Sum-Product Networks (SPNs, see, e.g., \cite{poon2011sum}), which are tree structures of arbitrary depth, with leaves that represent tractable distribution families (typically multi-variate normal, binomial or student-t distributions). 
SPN nodes perform either weighted summation or multiplication, and with appropriate constraints on the tree structure it can be shown that sampling, inference and density estimation remain tractable. 
Problems include finding a suitable SPN structure, and dealing with complex and high-dimensional data. 
Convolutional extensions of the SPN model exist, although they have been applied to very small images only \cite{butz2019deep}.
A similar approach is represented by probabilistic circuits (PCs, see, e.g., \cite{peharz2020einsum}) or Tensorial Mixture Models (TMMs, see, e.g., \cite{sharir2016tensorial}).
\smallskip\par
\noindent\textbf{Convolutional \acp{GMM}}\hspace{.5em}
The only work we could identify proposing hierarchical convolutional \acp{GMM} is \cite{GhazvinianZanjani2018}.
This work describes a hybrid model combining a \ac{CNN} and a \ac{GMM}.
\smallskip\par
\noindent\textbf{\acs{SGD} and End-To-End Training for GMMs}\hspace{.5em}
Training \acp{GMM} by \ac{SGD} is challenging due to local optima and the need to enforce model constraints, most notably the constraint of positive-definite covariance matrices.
At the same time, SGD is attractive in deep architectures due to its simplicity, which is facilitated even further by modern machine learning packages that perform automatic differentiation.
SGD for GMMs has recently been discussed in \cite{Hosseini2020}, although the proposed solution requires parameter initialization by \mbox{k-means} and introduces several new hyper-parameters.
Thus, it is unlikely to work as-is in a hierarchical structure.
An \ac{SGD} approach that achieves robust convergence even without k-means-based parameter initialization is presented by \cite{gepperth2020gradientbased}.
Undesirable local optima caused by random parameter initialization are circumvented by an adaptive annealing strategy.
SPNs and PCs can be trained end-to-end by SGD as well, although no article describes this in detail.
It is presumable that data-driven initialization is required, as well.
Previous hierarchical GMM proposals \cite{Viroli2019,VanDenOord2014,Tang2012} use (quite complex) extensions of the \ac{EM} algorithm initialized by k-means for training. 
In \cite{Richardson2018}, training is performed using \ac{SGD}, although with a k-means initialization.
\section{DCGMM: Model Overview}\label{sec:over}
\noindent The \acf{DCGMM} is a hierarchical model consisting of \textit{layers} in analogy to \acp{CNN}.\footnote{Repository: \url{https://gitlab.cs.hs-fulda.de/ML-Projects/dcgmm}} 
Each layer with index $L$ expects an input tensor $\tA^{(L-1)}$\,$\in$\,$\mathbb{R}^4$ of the dimensions $N,H^{(L-1)},W^{(L-1)},C^{(L-1)}$ and produces an output tensor $\tA^{(L)}$\,$\in$\,$\mathbb{R}^4$ of dimensions the $N,H^{(L)},W^{(L)},C^{(L)}$.
Layers can have internal variables $\vtheta^{(L)}$ that are adapted during \ac{SGD} training.
\par
\begin{figure*}[h]
	\centering
	\includegraphics[width=.8\textwidth]{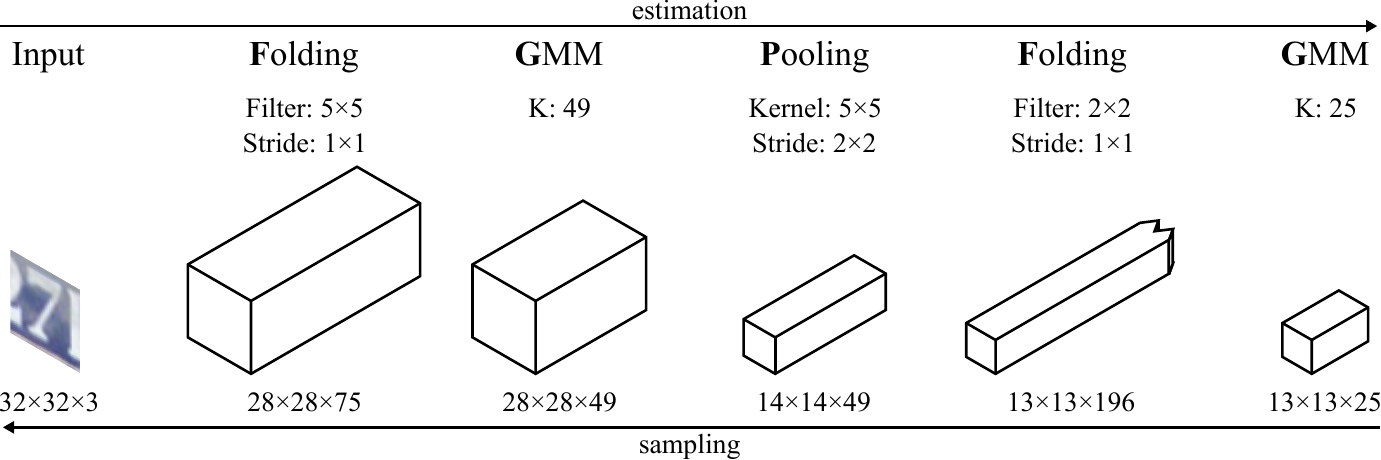} 
	\caption{Illustration of a \Ac{DCGMM} instance, with exemplary dimensionalities and parameters for each layer.\label{fig:global}}
\end{figure*}

A \ac{DCGMM} layer $L$ has two basic operating modes (see \cref{fig:global}): for (density) \textit{estimation}, an input tensor $\tA^{(L-1)}$ from layer $L$\,$-$\,$1$ is transformed into an output tensor $\tA^{(L)}$. 
For \textit{sampling}, the direction is reversed: each layer receives a control signal $\tT^{(L+1)}$ from layer $L$\,$+$\,$1$ (same dimensions as $\tA^{(L)}$), which is transformed into a control signal $\tT^{(L)}$ to layer $L$\,$-$\,$1$ (same dimensions as $\tA^{(L-1)}$).
\subsection{Layer Types}\label{sec:layers}
\noindent We define four layer types: \ac{F}, \ac{P}, \ac{C} and \ac{G}. 
Each implements distinct operations for both modes, i.e., estimation and sampling.
\subsubsection{\acl{F} Layer}
For density estimation, this layer performs a part of the convolution operation well-known from \acp{CNN}.
Based on the filter sizes $f_X^{(L)}$, $f_Y^{(L)}$ and the filter strides $\Delta_X^{(L)}$, $\Delta_Y^{(L)}$, all entries of the input tensor inside the range of the sliding filter window are \enquote{dumped} into the channel dimension of the output tensor.
We thus obtain an output tensor of dimensions $N,H^{(L)}$\,$=$\,$1$\,$+$\,$\frac{H^{(L-1)}\!-\!f_Y^{(L)}}{\Delta_Y^{(L)}}$, $W^{(L)}$\,$=$\,$1$\,$+$\,$\frac{W^{(L-1)}\!-\!f_X^{(L)}}{\Delta_X^{(L)}}$ and $C^{(L)}$\,$=$\,$C^{(L)}f_X^{(L)}f_Y^{(L)}$, whose entries are computed as $\etA^{(L)}_{nhwc}$\,$=$\,$\etA^{(L-1)}_{nh'w'c'}$ with $h$\,$=$\,$h'/f_Y^{(L)}$, $w$\,$=$\,$w'/f_X^{(L)}$ and $c$\,$=$\,$c'$\,$+$\,$\big((h'$\,$-$\,$h\Delta_Y^{(L)})f_X^{(L)}$\,$+$\,$w'$\,$-$\,$w\Delta_X^{(L)}\big)C^{(L-1)}$\,$+$\,$c'$.
When sampling, the inverse mapping is performed which is not a one-to-one correspondence: input tensor elements which receive several contributions are averaged over all these.
\subsubsection{\acl{P} Layer}
For density estimation, pooling layers perform the same operations as standard (max-)pooling layers in \acp{CNN} based on the kernel sizes $k_Y^{(L)}$, $k_X^{(L)}$ and strides $\Delta_X^{(L)}$, $\Delta_Y^{(L)}$. 
When sampling, pooling layers perform a simple nearest-neighbor up-sampling by a factor indicated by the kernel sizes and strides.
\subsubsection{\acl{G} Layer}\label{sec:glayer} 
This layer type contains $K$ \ac{GMM} components, each of which is associated with trainable parameters $\pi_k$, $\vmu_k$ and $\mSigma_k$, $k$\,$=$\,$1\dots K$, representing the \ac{GMM} \textit{weights}, \textit{centroids} and \textit{covariances}. 
What makes \acl{G} layers convolutional is that they do not model single input vectors, but the channel content at \textit{all} positions $h,w$ of the input $\tA^{(L-1)}_{n,w,h,:}$, using a shared set of parameters.
This is analogue to the way a CNN layer models image content at all sliding window positions using the same filters.
A \acl{G} layer maps the input tensor $\tA^{(L-1)}$\,$\in$\,$\mathbb{R}^{N,H^{(L-1)},W^{(L-1)},C^{(L-1)}}$ to $\tA^{(L)}$\,$\in$\,$\mathbb{R}^{N,H^{(L-1)},W^{(L-1)},K}$, each \ac{GMM} component $k$\,$\in$\,$\{1,\dots,K\}$ contributing the likelihood $\etA^{(L)}_{nhwk}$ of having generated the channel content at position $h,w$ (for sample $n$ in the mini-batch).
This likelihood is often referred to as \textit{responsibility} and is computed as
\begin{equation}
	\begin{split}
		p_{nhwk} \big(\tA^{(L-1)}\big) & = \mathcal{N}_k\big(\tA^{(L-1)}_{nhw:}; \vmu_k,\mSigma_k\big) \\
		\etA^{(L)}_{nhwk}              & \equiv  \frac{p_{nhwk}}{\sum_{c'} p_{nhwc'}}.
	\end{split}
	\label{equ:responsibilities}
\end{equation}
For training the \acl{G} layer, we optimize the \ac{GMM} log-likelihood $\mathcal{L}^{(L)}$ for each layer $L$: 
\begin{equation}
	\begin{split}
		\mathcal{L}^{(L)}_{hw} & = \sum_n \log \sum_k \pi_k p_{nhwk}(\tA^{(L-1)})             \\
		\mathcal{L}^{(L)}      & = \frac{\sum_{hw}\mathcal{L}^{(L)}_{hw}}{H^{(L-1)}W^{(L-1)}}
	\end{split}
	\label{eqn:loglik}
\end{equation}
Training is performed by \ac{SGD} according to the technique, and with the recommended parameters, presented by \cite{gepperth2020gradientbased}, which uses a max-component approximation to $\mathcal{L}^{(L)}$.
In sampling mode, a control signal $\tT^{(L)}$ is produced by standard \ac{GMM} sampling, performed separately for all positions $h,w$.
\ac{GMM} sampling at position $h,w$ first selects a component by drawing from a multinomial distribution.
If the \acl{G} layer is the last layer of a \ac{DCGMM} instance, the multinomial's parameters are the mixing weights $\pi_{:}$ for each position $h,w$.
Otherwise, the control signal $\tT^{(L+1)}_{nhw:}$ received from layer $L$\,$+$\,$1$ is used.
It is consistent to use the control signal for component selection in layer $L$, since it was sampled by layer $L$\,$+$\,$1$, which was in turn trained on the component responsibilities of layer $L$, see \cref{sec:diff}.
The selected component (still at position $h,w$) then samples $\tT^{(L)}_{nhw:}$.
It is often beneficial for sampling to restrict the selected number to the $S$ components with the highest control signal (top-$S$ sampling).
This reduces the diversity of sampling as fewer components can participate, but improves its quality since selection is restricted to the most likely components.
\subsubsection{\acl{C} Layer}\label{sec:clayer}
This layer type implements a linear classifier, trained in a supervised manner by cross-entropy loss.
Logits are obtained via an affine transformation $\tA^{(L)}$\,$=$\,$\hat \tA^{(L-1)}\mW^{(L)}$\,$+$\,$\vb^{(L)}$ from the flattened input activities $\hat \tA^{(L-1)}$\,$\in$\,$\mathbb{R}^{N\times HWC}$. 
The logits have the dimension $N$\,$\times$\,$K$, with $K$ representing the number of classes or categories.
For sampling, a control signal is generated by approximately inversing this transformation: $\tT^{(L-1)}$\,$=$\,$\mW^{(L),T}\tT^{(L)}$\,$-$\,$\vb^{(L)}$, where $\tT^{(L)}$\,$\in$\,$\mathbb{R}^{N\times K}$ contains the one-hot coded classes for each sample to be generated.
\subsection{Architecture-Level Functionalities}\label{sec:arch}
\subsubsection{End-to-End Training}\label{sec:end2end}
To train an \ac{DCGMM} instance, we optimize $\mathcal{L}^{(L)}$ for each \acl{G} layer $L$ by vanilla SGD\footnote{Advanced \ac{SGD} strategies like RMSProp \cite{Hinton2012} or Adam \cite{Kingma2015} seem incompatible with \ac{GMM} optimization.}, using learning rates $\epsilon^{(L)}$.
This is different from a standard \ac{CNN} classifier, where only a single loss function is minimized.
Usually a cross-entropy loss is computed from the last layer's outputs. 
Learning is \textit{not} conducted layer-wise but end-to-end.
Parameter initialization for \acl{G} layers selects the initial values for the mixing weights as $\pi_{k}$\,$=$\,$K^{-1}$.
Centroid elements sampled from $\vmu_{kl}$\,$\sim$\,$\mathcal{U}_{[\text{-}0.01,0.01]}$ and diagonal covariances are initialized to unit entries.
Training is conducted in two steps to ensure convergence.
First, only centroids are adapted, whereas both centroids and precisions are adapted in the second step.

\subsubsection{Density Estimation and Outlier Detection}\label{sec:methods:outlier}
Outlier detection requires the computation of long-term averages in all layers and positions, $\E_n \mathcal{L}^{(L)}_{nhw}$ and variances $\text{Var}_n(\mathcal{L}^{(L)}_{nhw})$ over the training set, preferably during a later, stable part of training. 
Thus, for every layer and position $h,w$, inliers are characterized by 
\begin{align}
	\label{inlier}
	\mathcal{L}^{(L)}_{hw} \ge \etB^{(L)}_{hw} \equiv \E_n \mathcal{L}^{(L)}_{nhw} - c\sqrt{\text{Var}_n(\mathcal{L}^{(L)}_{nhw})}.
\end{align}
A larger $c$ implies a less restrictive identification of inliers.
\par
Assuming that the topmost \acl{G} layer is global ($h$\,$=$\,$w$\,$=$\,$1$), \cref{inlier} reduces to a single condition that determines whether the sample as a whole is an inlier.
However, we can also localize inlier/outlier image parts by evaluating \cref{inlier} in lower \acl{G} layers.
\subsubsection{Sampling and Sharpening}\label{sec:sampling}
Sampling starts in the highest layer $L$, assumed to be an \acl{G} layer, and propagates control signals downwards (see \cref{fig:global} and \cref{sec:glayer}), with control signal $\tT^{(L)}$ constituting the sampling result.
Sampling suffers from information loss due to the non-invertible mappings effected by \acl{P} and \acl{F} layers.
To counteract this, a \acl{F} or \acl{P} layer at $L-1$ performs \textit{sharpening} on the control signal $\tT^{(L-1)}$ it generates from $\tT^{(L)}$.
This involves computing $\mathcal{L}^{(L)}(\tT^{(L-1)})$ for the \acl{G} layer at level $L$ and performing $G$ gradient ascent steps $\tT^{(L-1)}_{nhwc}$\,$\rightarrow$\,$\tT^{(L-1)}_{nhwc}$\,$+$\,$\epsilon_{s} \partial \mathcal{L}^{(L)} / \partial \tT^{(L-1)}_{nhwc}$. 
The reason for sharpening is that filters in \acl{F} layers usually overlap, and neighboring filter results are correlated.
This correlation is captured by all higher \acl{G} layers, and most prominently by the next-highest one.
Therefore, modifying $\tT^{(L-1)}$ by gradient ascent will recover some of the information lost by pooling or folding.
After sharpening, the tensor $\tT^{(L-1)}$ is passed as control signal to $L$\,$-$\,$2$.
\subsubsection{Conditional Sampling} 
By conditional sampling we understand the ability to selectively produce samples from a certain class.
Obviously, this is only possible when the DCGMM instance has a top-level linear classifier layer.
We proceed as in \cref{sec:sampling}, except for the component selection in the top-level \acl{G} layer.
For conditional sampling, we simply train a linear classifier in the last layer $L$ on the outputs $\tA^{(L-1)}$ of the previous layer, and then approximately invert this mapping given a certain class label to obtain a control signal $\tT^{(L)}$ for sampling in layer $L$\,$-$\,$1$.
\subsubsection{In-Painting} 
This is a functionality where a corrupted image, from which a part has been deleted, is supplied to a trained \ac{DCGMM} instance, which then \textit{completes} the missing parts. 
Here, we add a twist by not informing the model on which parts of the image have been deleted.
Rather, a DCGMM should infer this on its own using its outlier detection capability, see \cref{sec:methods:outlier}. 
\par
Mathematically, in-painting requires that we perform \textit{posterior inference} from a trained DCGMM.
We shall consider this kind of inference for a \enquote {flat} GMM first, and subsequently generalize to DCGMMs.
\par
Assuming that an image $\vx$ is composed of two parts $\vx_1$ and $\vx_2$, where $\vx_2$ is corrupted. To recover it, we wish to draw samples from the distribution $p(\vx_2|\vx_1)$.
A simple computation yield an expression  that can be evaluated by re-introducing the latent variable $z$ of the GMM, see \cref{eqn:prob:complete}:
\begin{equation}
	\begin{split}
	p(\vx_2|\vx_1) & = \frac{p(\vx_1\vx_2)}{p(\vx_1)} = \sum_z \frac{p(\vx_1,z)}{p(\vx_1,z)} \frac{p(\vx_2\vx_1,z)}{p(\vx_1)} = \\
	               & = \sum_z p(\vx_2|\vx_1,z) p(z|\vx_1) \equiv \sum_z \gamma_z p(\vx_2|\vx_1,z).
\end{split}
\raisetag{35pt}
\end{equation}
This shows that the distribution we wish to sample from is again a Gaussian mixture.
The mixture coefficients are given by the \textit{responsibilities} $\gamma_z = p(z|\vx_1)$, whereas the component densities can be simplified as $p(\vx_2|\vx_1,z) \sim p(\vx_2\vx_1,z)$. 
We first draw a value $z^*$ from a multi-nomial distribution defined by $\gamma_i$, and then generate a sample $\vx$ from component probability $p(\vx_2\vx_1,z^*)$ of the Gaussian mixture. 
The part of $\vx$ that belongs to $\vx_1$ is simply limited when sampling. 
\par
In a DCGMM, we implement this procedure for every position $(h,w)$ in all GMM layers.
We start at the top of the hierarchy (layer $O$), where responsibilities are simply the activities $\tA^{(O)}$ computed by forward-propagating the corrupted input in estimation mode.
The responsibilities are used to draw $z^*(h,w)$ for each position $(h,w)$ as detailed above.
Using $z^*$, we obtain a control signal for layer $O$\,$-$\,$1$ as
\begin{align}
\tT^{(O)}_{hw:} \sim \mathcal N_{z^*(h,w)}(\cdot), 
\end{align}
which propagated downwards through pooling and folding layers to the next-deeper GMM layer. 
\par
As for each GMM layer $L$\,$<$\,$O$, we want to in-paint only those parts which are not corrupted.
We combine the raw control signal $\tT^{(L)}$ and the activities $\tA^{(L)}$ obtained in the forward pass into a fused control signal
\begin{align}
\hat \tT^{(L)}_{nhw:} = \left\{
\begin{array}{cl}
	\tA^{(L)}_{nhw:} & \text{ if inlier at }h, w   \\
	\tT^{(L)}_{nhw:} & \text{ if outlier at }h, w
\end{array}\right.
\end{align}
where outliers are determined as described in \cref{sec:methods:outlier}.
Thus, control signals \enquote{fill in} the activities that seem corrupted.
\subsubsection{Variant Generation}
Variant generation is a special case of sampling, where a template image is supplied and the DCGMM generates similar ones. 
This requires forward-propagating the template image in estimation mode through the DCGMM, and subsequently sampling a control signal from the top-level GMM layer $O$ according to the obtained top-level activities $\tA_{hw:}^{(O)}$.
For all layers $L$\,$<$\,$O$ sampling is performed as detailed in \cref{sec:sampling}. 
An interesting option is to control the similarity to the template image.
This can be done either by selecting a suitable $S$ for top-$S$-sampling, or by limiting top-down signals to activities in higher DCGMM layers: 
\begin{align}
	\tT^{(L)} \equiv \tA^{(L-1)} \text{ for } L \ge \hat{L},
\end{align}
where $\hat{L}$ is a free parameter.
The closest match between template image and generated image is to be expected for $\hat{L}$\,$=$\,$0$, whereas $\hat{L}$\,$>$\,$O$ should result in very diverse results.
\section{Experiments}\label{sec:experiments}
\setlength\tabcolsep{2pt}
\begin{table*}[b!]
	\centering
	\caption{Configurations of different \ac{DCGMM} architectures.\label{tab:dcgmm_architectures}}
	\small 
	\begin{tabular}{|c|l|l|l|l|l|l|l|l|}
		\hline
		\diagbox[width=36.5pt,height=1.9em]{layer}{ID} & \cen{$1L$}    & \cen{$2L$-$a$} & \cen{$2L$-$b$} & \cen{$2L$-$c$} & \cen{$2L$-$d$}     & \cen{$2L$-$e$} & \cen{$3L$-$a$} & \cen{$3L$-$b$}     \\ \hline
		                      1                        & F(28,28,1,1)  & F(20,20,8,8)   & F(7,7,7,7)     & F(8,8,2,2)     & F(28,28,1,1)       & F(4,4,2,2)     & F(3,3,1,1)     & F(28,28,1,1)       \\
		                      2                        & G(25)         & G(25)          & G(25)          & G(25)          & G(25)              & G(25)          & G(25)          & G(25)              \\
		                      3                        &               & F(2,2,1,1)     & F(4,4,1,1)     & F(11,11,1,1)   & F(1,1,1,1)         & F(13,13,1,1)   & P(2,2)         & F(1,1,1,1)         \\
		                      4                        &               & G(36)          & G(36)          & G(36)          & G(36)              & G(36)          & F(4,4,1,1)     & G(25)              \\
		                      5                        &               &                &                &                &                    &                & G(25)          & F(1,1,1,1)         \\
		                      6                        &               &                &                &                &                    &                & P(2,2)         & G(25)              \\
		                      7                        &               &                &                &                &                    &                & F(6,6,1,1)     &                    \\
		                      8                        &               &                &                &                &                    &                & G(49)          &                    \\ \hline
		               \textit{comment}                & \cen{vanilla} & \cen{1 conv.}  & \cen{1 conv.}  & \cen{1 conv.}  & \cen{no}           & \cen{1 conv.}  & \cen{2 conv.}  & \cen{no}           \\
		                                               & \cen{GMM}     & \cen{layer}    & \cen{layer}    & \cen{layer}    & \cen{convolutions} & \cen{layer}    & \cen{layers}   & \cen{convolutions} \\ \hline
	\end{tabular}
\end{table*}
\noindent We define various \ac{DCGMM} instances (with 2 or 3 GMM layers) for evaluation, see \cref{tab:dcgmm_architectures}, plus a single-layer \ac{DCGMM} baseline which is nothing but a vanilla GMM.
A \ac{DCGMM} instance is defined by the parameters of its layers:
\textbf{\ac{F}}olding($f_Y$,\,$f_X$,\,$\Delta_Y$,\,$\Delta_X$), (Max-)\textbf{\ac{P}}ooling($k_Y$,\,$k_X$,\,$\Delta_Y$,\,$\Delta_X$) and \textbf{\ac{G}}MM($K$). 
Unless stated otherwise, training is always conducted for 25 epochs, using the recommended parameters from \cite{gepperth2020gradientbased}.
Sharpening is always performed for $G$\,$=$\,$1\,000$ iterations with a step size of $0.1$.
\subsection{Datasets}
\noindent For the evaluation we use the following image datasets:
\smallskip\\
\noindent\textbf{MNIST}~\cite{LeCun1998} is the common benchmark for computer vision systems and classification problems.
It consists of $60\,000$ $28$\,$\times$\,$28$ gray scale images of handwritten digits (0-9).
\smallskip\\
\textbf{FashionMNIST}~\cite{Xiao2017} consists of images of clothes in 10 categories and is structured like the MNIST dataset.
\par\smallskip
Although these datasets are not particularly challenging for classification, their dimensionality of $784$ is at least one magnitude higher than datasets in the literature, which are used for validating other hierarchical GMM approaches in the literature.
\subsection{Sampling, Sparsity and Interpretability}
\begin{figure*}[t]
	\centering
	\includegraphics[width=.9\linewidth]{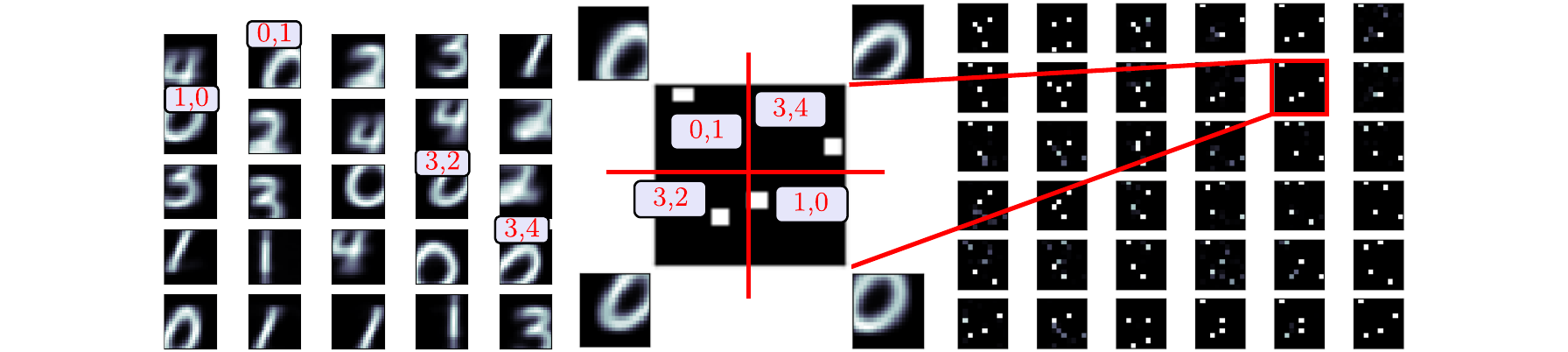}
	\caption{\label{fig:sparse} 
		Sampling from \ac{DCGMM} instance $2L$-$a$, see \cref{tab:dcgmm_architectures}. 
		Learned \ac{GMM} centroids (left: G2, right: G4) are presented along with an illustration of sampling, having initially selected the layer $4$ component highlighted in red. 
		In the middle, the selected G4 centroid is shown.
	}
\end{figure*}
\noindent We show that trained \ac{DCGMM} parameters are sparse and have an intuitive interpretation in terms of sampling.
To this effect, we train \ac{DCGMM} instance $2L$-$a$ (see \cref{tab:dcgmm_architectures}). 
After training (see \cref{sec:end2end}), we plot and interpret the centroids of the \acl{G} layers $2$ (G2) and $4$ (G4). 
The centroids of layer 2 (left of \cref{fig:sparse}) are easily interpretable and reflect the patterns that can occur in any of the $2$\,$\times$\,$2$ input patches to layer G2 of size $20$\,$\times$\,$20$.
The $36$\,$=$\,$6$\,$\times$\,$6$ centroids of G4 (right of \cref{fig:sparse}) express typical responsibility patterns computed from each of the $2$\,$\times$\,$2$ input patches to G2.
They are, very sparsely populated.
Another interpretation of G4 centroids can be found in terms of sampling (see \cref{sec:sampling}), which would first select a random G4 component to generate a sample of dimensions $H,W,C$\,$=$\,$1,1,2$\,$\times$\,$2$\,$\times$\,$5$\,$\times$\,$5$, and pass it on as a control signal to G2.
Traversing Folding layer 3 only reshapes the control signal to dimension $H,W,C$\,$=$\,$2,2,5$\,$\times$\,$5$, depicted in the middle of \cref{fig:sparse}. 
This signal controls component selection in each of the $2$\,$\times$\,$2$ positions in G2.
Due to their sparsity, we can directly read off the components likely to be selected for sampling at each position.
Thus, G2 generates a control signal whose $2$\,$\times$\,$2$ positions of dimensions $H,W,C$\,$=$\,$20,20,1$ overlap in the input plane (this is resolved by sharpening in Folding layer 1).
In this case, it is easy to see that sampling produces a particular representation of the digit zero.
\subsection{Outlier Detection}
\begin{figure}[htb!]
	\centering
	\includegraphics[width=.7\linewidth]{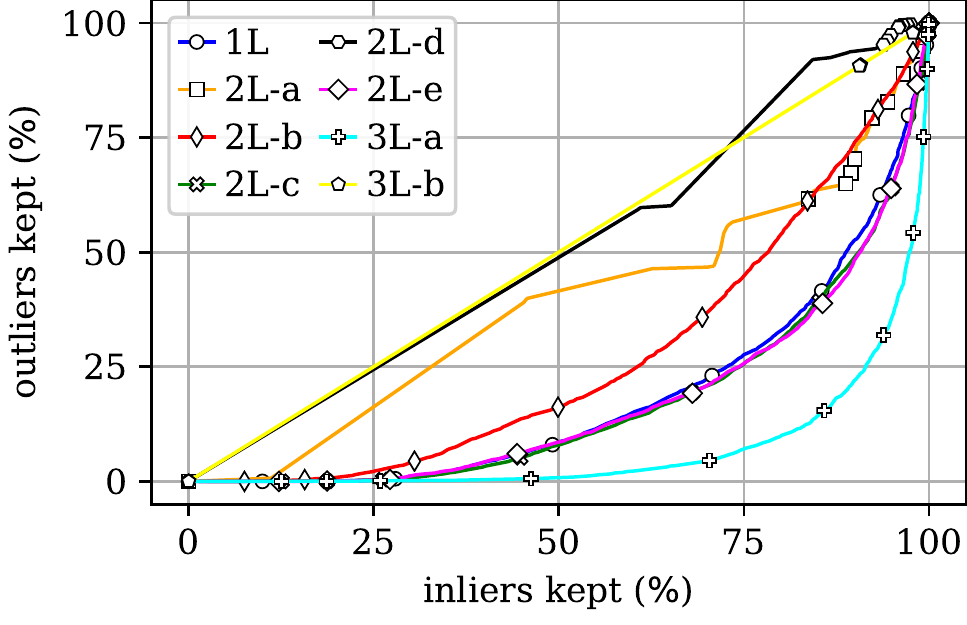}\\
	\includegraphics[width=.7\linewidth]{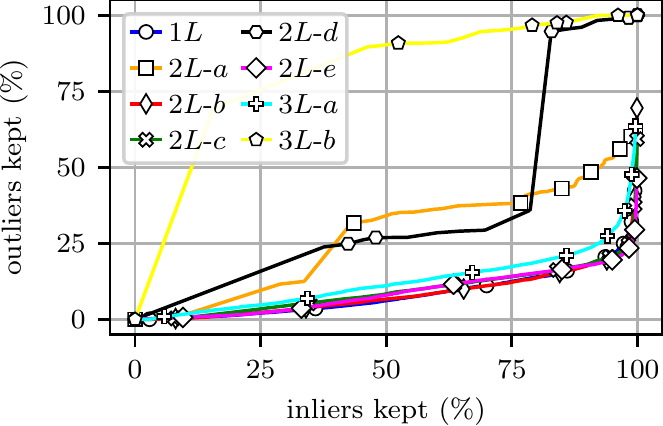} 
	\caption{Visualization of outlier detection capabilities (different \acp{DCGMM}) for MNIST (above) and FashionMNIST (below).}
	\label{fig:out1}
\end{figure}

\noindent For outlier detection, we compare \ac{DCGMM} architectures from \cref{tab:dcgmm_architectures}, using the log-likelihood of the highest layer as a criterion as detailed in \cref{sec:methods:outlier}.
We first train a \ac{DCGMM} instance on classes 0-4, and subsequently use the trained classes for inlier- and class 5-9 for outlier-detection.
We vary $c$ in the range $[-2,2]$, resulting in different outlier and inlier percentages. 
\cref{fig:out1} shows the ROC-like curves which clearly indicate that the deep convolutional \Ac{DCGMM} instances perform best.
However, deep but non-convolutional instances like $2L$-$d$ and $3L$-$b$ consistently perform badly. 
\subsection{Clustering}
\noindent We compare \acp{DCGMM} to vanilla \acp{GMM} using established clustering metrics, namely the Dunn index~\cite{Dunn1973}  and the Davies-Bouldin score~\cite{Davies1979}.
The \ac{DCGMM} instances from \cref{tab:dcgmm_architectures} are tested on both image datasets.
We observe (see \cref{tab:out2} and \cref{fig:out1}) that mainly the deep but non-convolutional \ac{DCGMM} instances perform well in clustering, whereas convolutional instances, even if they are deep, are compromised.
These metrics do not measure the classification accuracy obtained by clustering but intrinsic clustering-related properties.
\begin{table}[h!]
	\centering
	\caption{ 
		Dunn index (higher is better) and Davies-Bouldin (DB) score (smaller is better), evaluated for all tested \ac{DCGMM} architectures on MNIST and FashionMNIST.
		Best are marked bold for $10$ repetitions (worst case).
	} 
	\label{tab:out2}
	\begin{tabular}{|lc|cccccccc|}
		\hline
		\multicolumn{1}{|l|}{\textbf{Dataset}} & \diagbox[width=50pt,height=2em]{Metric}{{DCGMM}} & $1L$ & $2L$-$a$ &   $2L$-$b$    & $2L$-$c$ &   $2L$-$d$    & $2L$-$e$ & $3L$-$a$ &   $3L$-$b$    \\ \hline
		\multirow{2}{*}{MNIST}                 &                    Dunn index                    & 0.14 &   0.14   &     0.13      &   0.12   & \textbf{0.19} &          &   0.15   &     0.15      \\ \cline{2-10}
		                                       &                     DB score                     & 2.59 &   2.73   &     3.06      &   2.62   &     2.57      &          &   2.65   & \textbf{2.53} \\ \hhline{----------}\hhline{----------}
		Fashion-                               &                    Dunn index                    & 0.14 &   0.15   & \textbf{0.16} &   0.15   &     0.11      &   0.11   &  0.096   &     0.13      \\ \cline{2-10}
		MNIST                                  &                     DB score                     & 2.37 &   2.77   &     2.62      &   2.7    &     2.40      &   2.92   &   3.2    & \textbf{2.35} \\ \hline
	\end{tabular}
\end{table}
\subsection{Sampling and Sharpening}
\noindent The results presented here were obtained by training on classes 0-4 of both datasets, and have to be confirmed by visual inspection of generated samples. 
The restriction to classes 0-4 is purely for visualization purposes.

\begin{figure}[htb]
	\centering
	\includegraphics[width=0.49\linewidth]{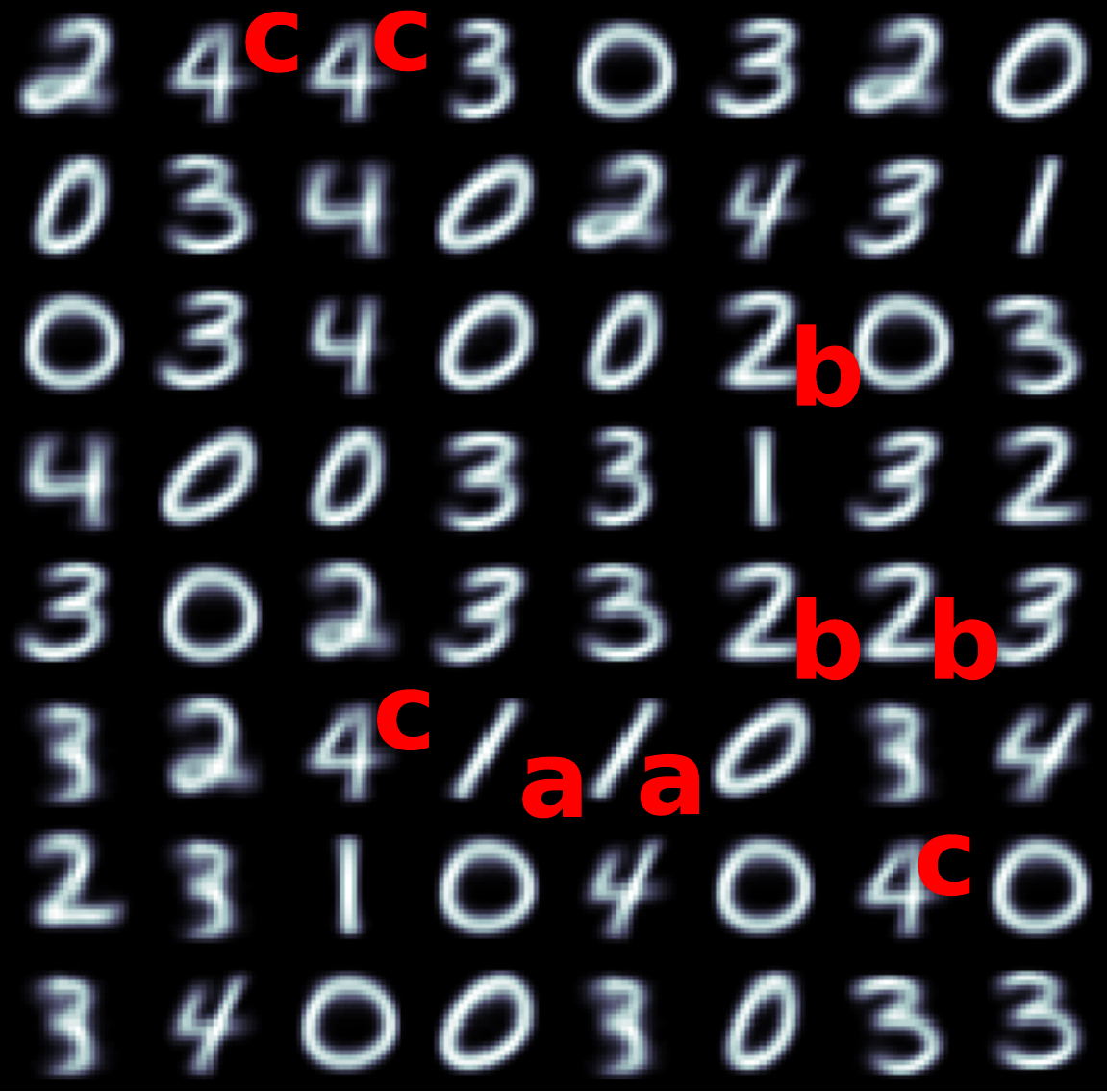}
	\includegraphics[width=0.49\linewidth]{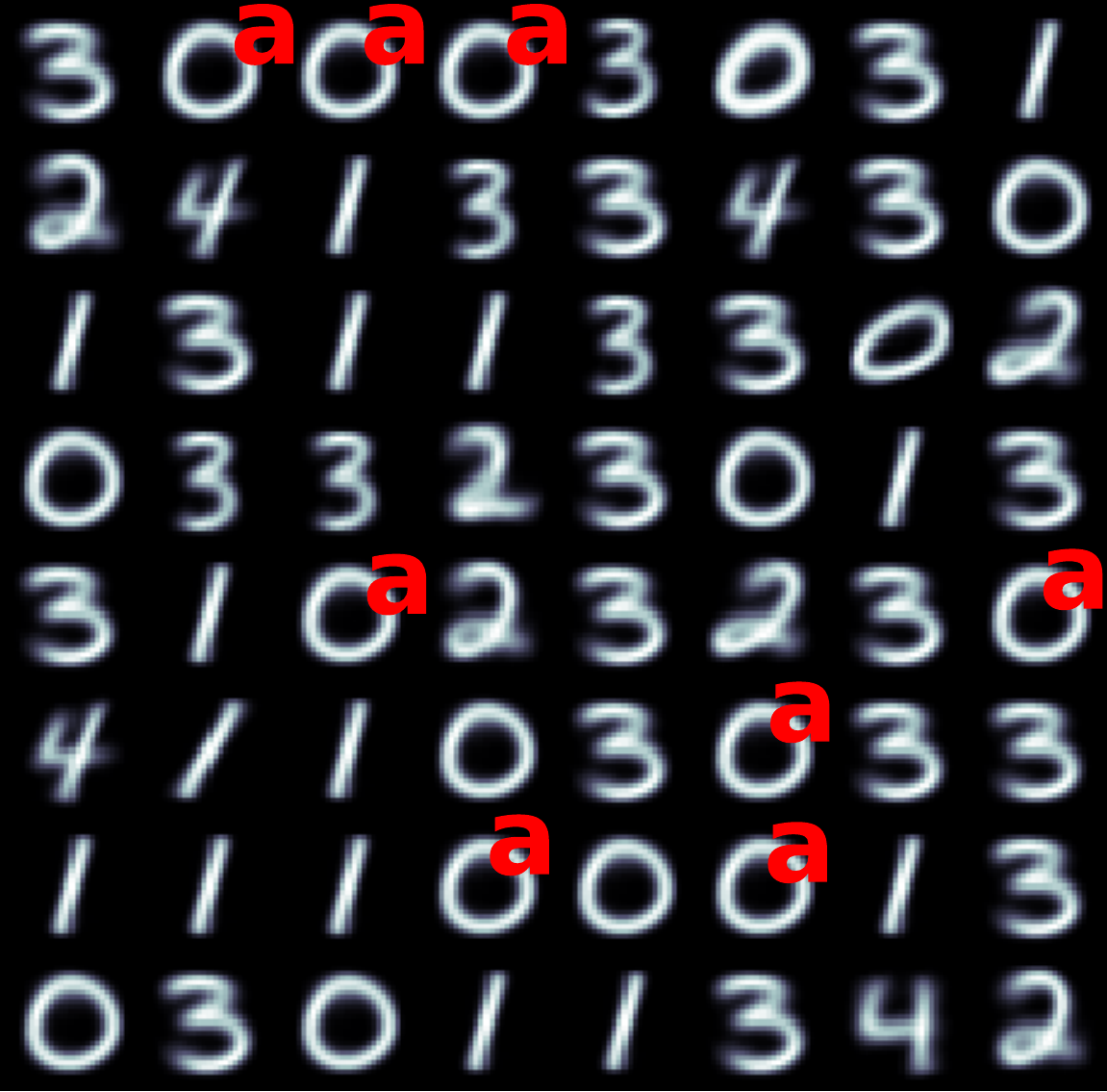}\\\vspace{.5em}
	\includegraphics[width=0.49\linewidth]{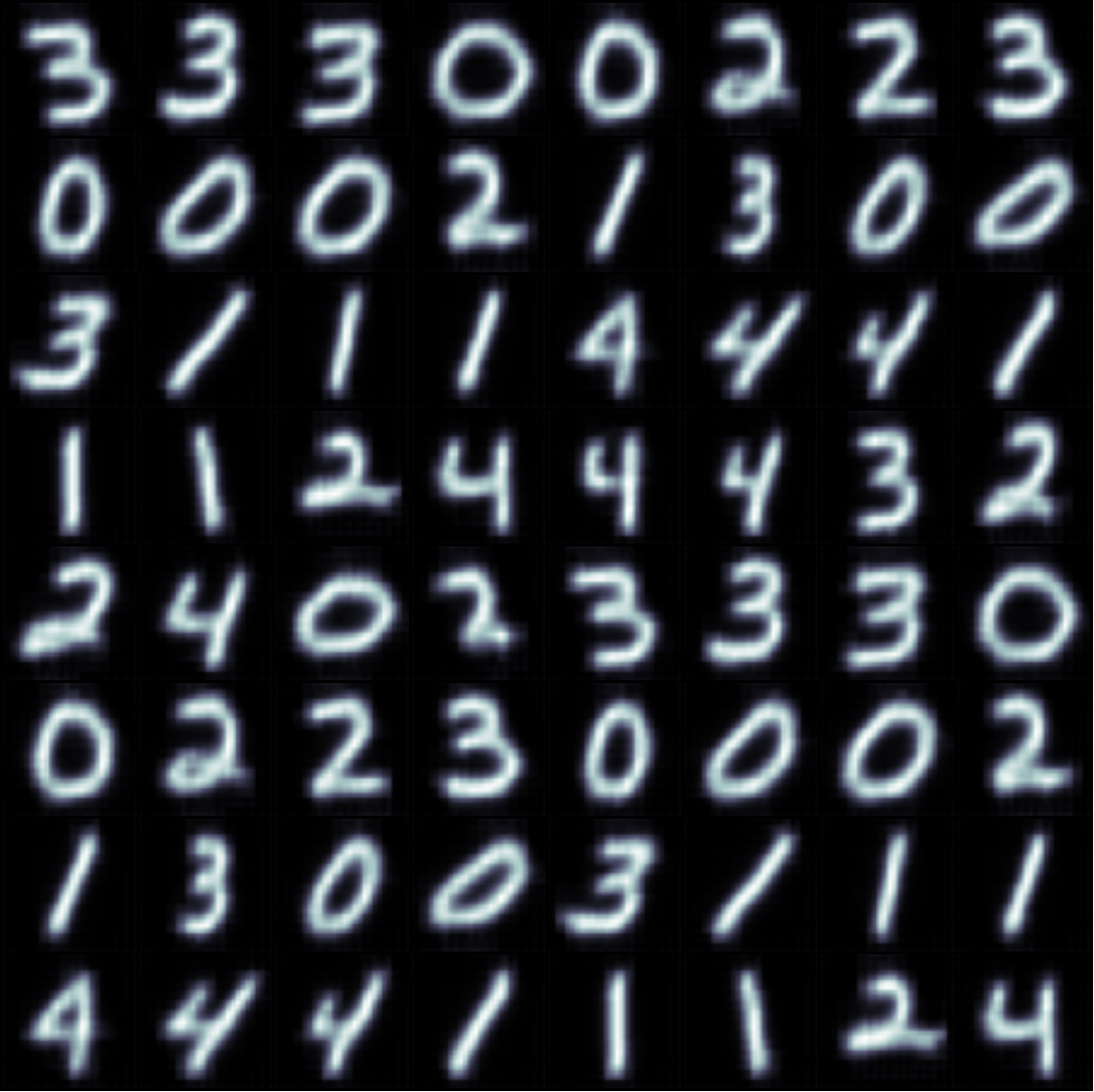}
	\includegraphics[width=0.49\linewidth]{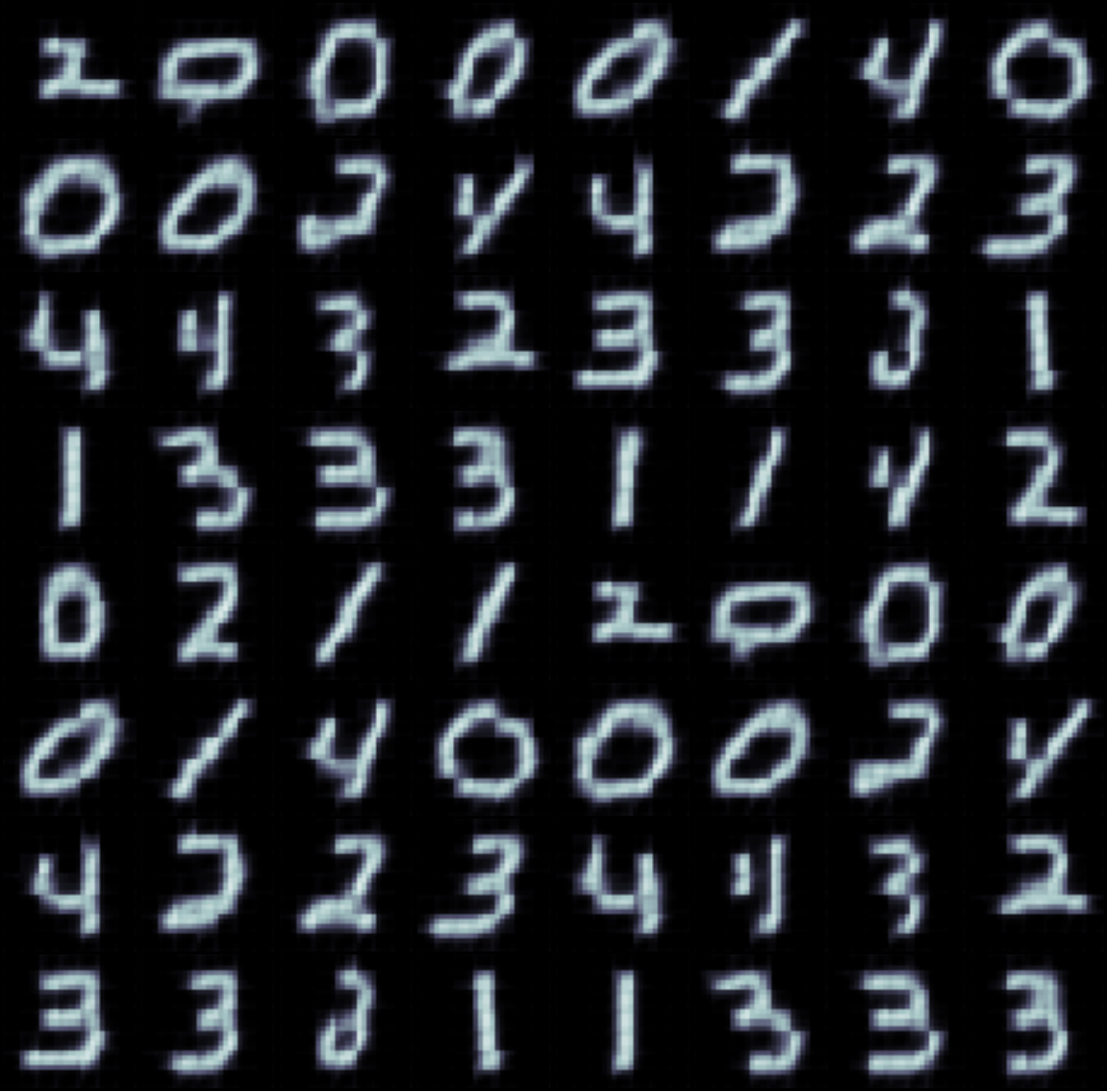}
	\caption{
		Convolutional architecture is helpful for sampling: top-1 sampling shown on MNIST for \ac{DCGMM} architectures $1L$ (vanilla GMM, upper left), $2L$-$d$ (non-convolutional 2-layer, upper right), $2L$-$c$ (convolutional 2-layer, lower left) and $2L$-$e$ (convolutional 2-layer, lower right).
		Please observe duplicated samples in the non-convolutional architectures, marked by identical red letters. 
		This figure is larger so duplicated samples can be better observed.
	}
	\label{fig:global_local}
\end{figure}

\begin{figure}[htb]
	\centering
	\begin{subfigure}[c]{0.25\linewidth}\includegraphics[width=\linewidth,trim=2.1cm 0cm 2.1cm 0cm,clip]{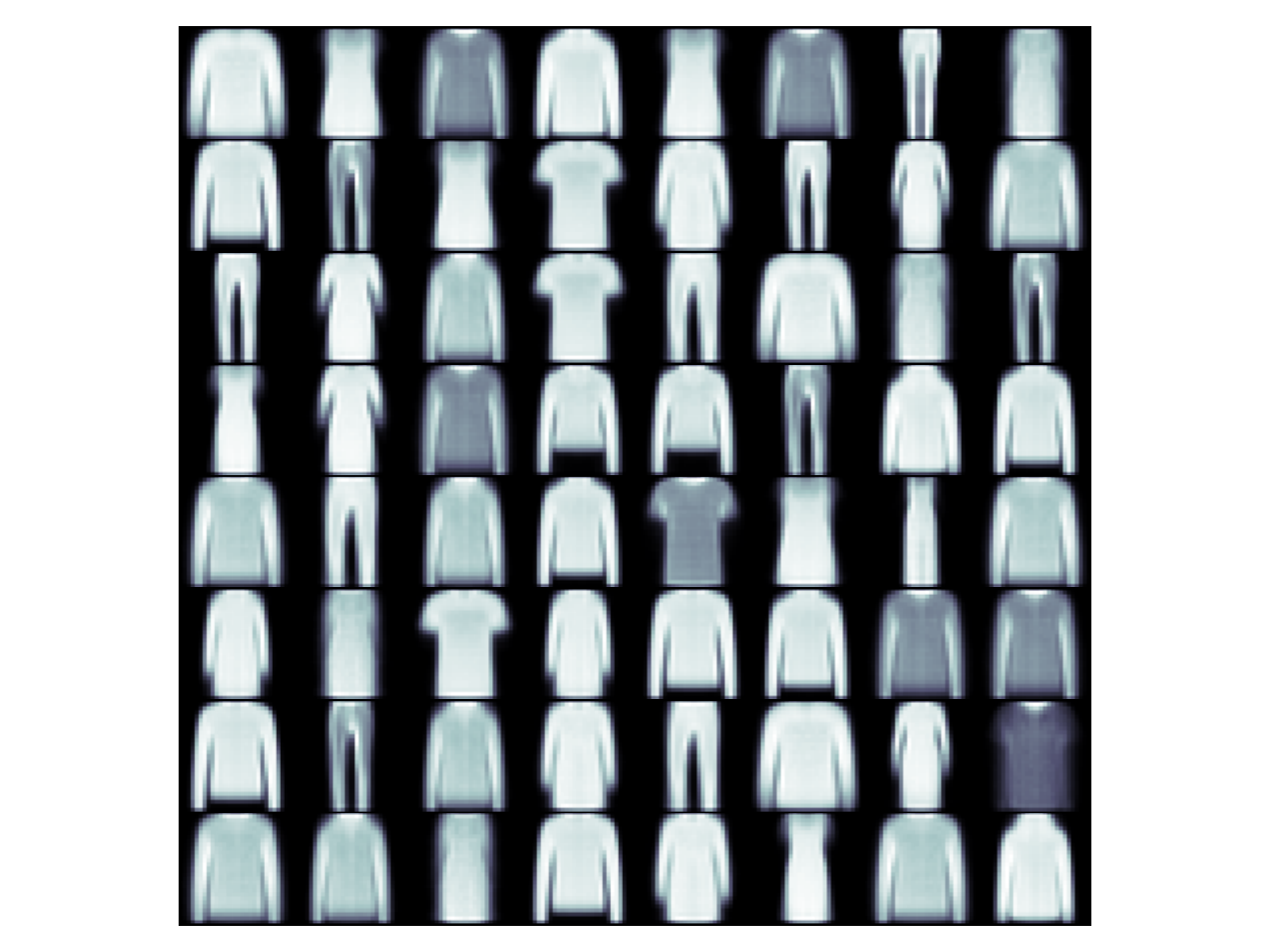}\subcaption{}\end{subfigure}\begin{subfigure}[c]{0.25\linewidth}\includegraphics[width=\linewidth,trim=2.1cm 0cm 2.1cm 0cm,clip]{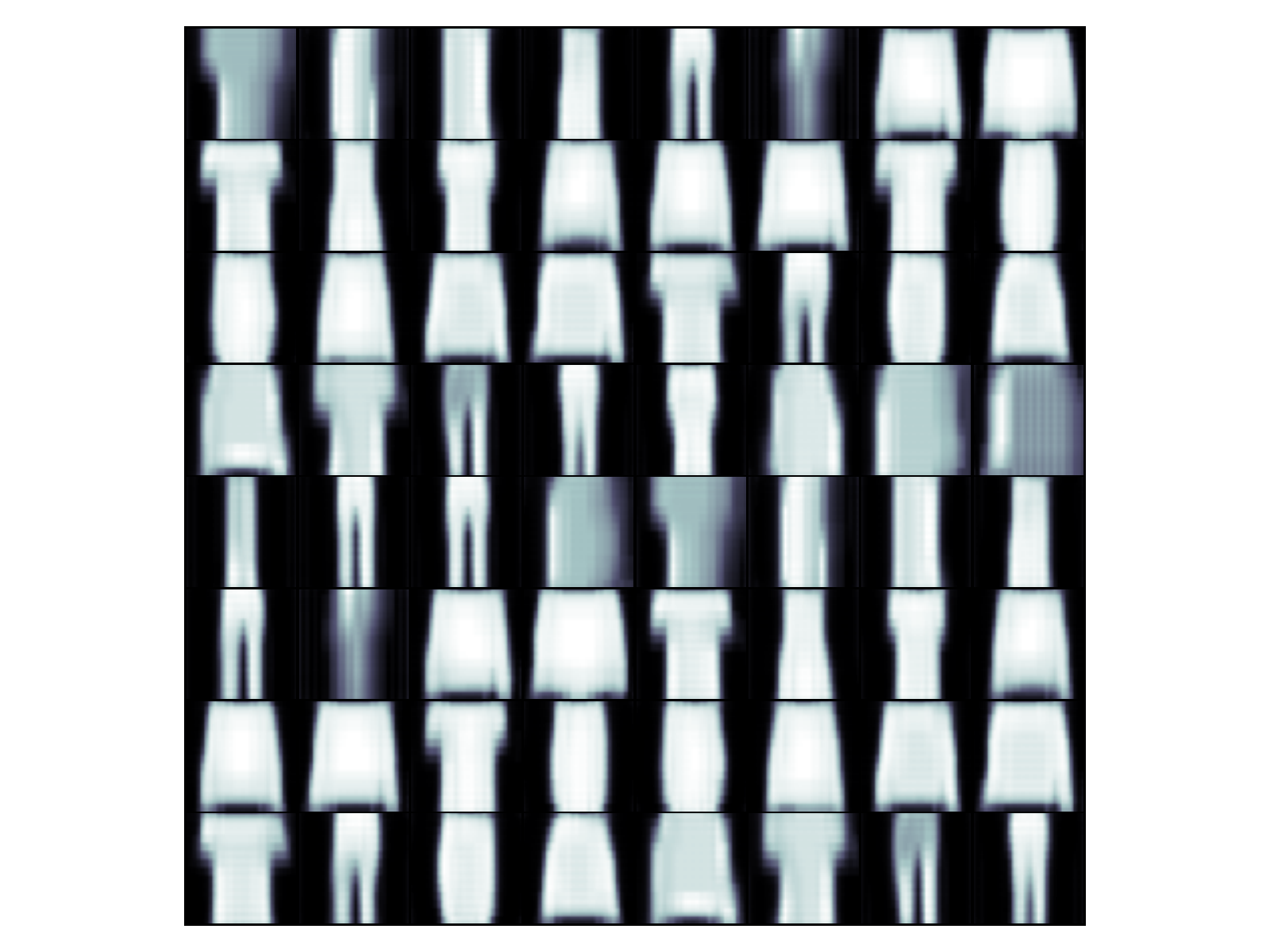}\subcaption{}\end{subfigure}\begin{subfigure}[c]{0.25\linewidth}\includegraphics[width=\linewidth,trim=2.1cm 0cm 2.1cm 0cm,clip]{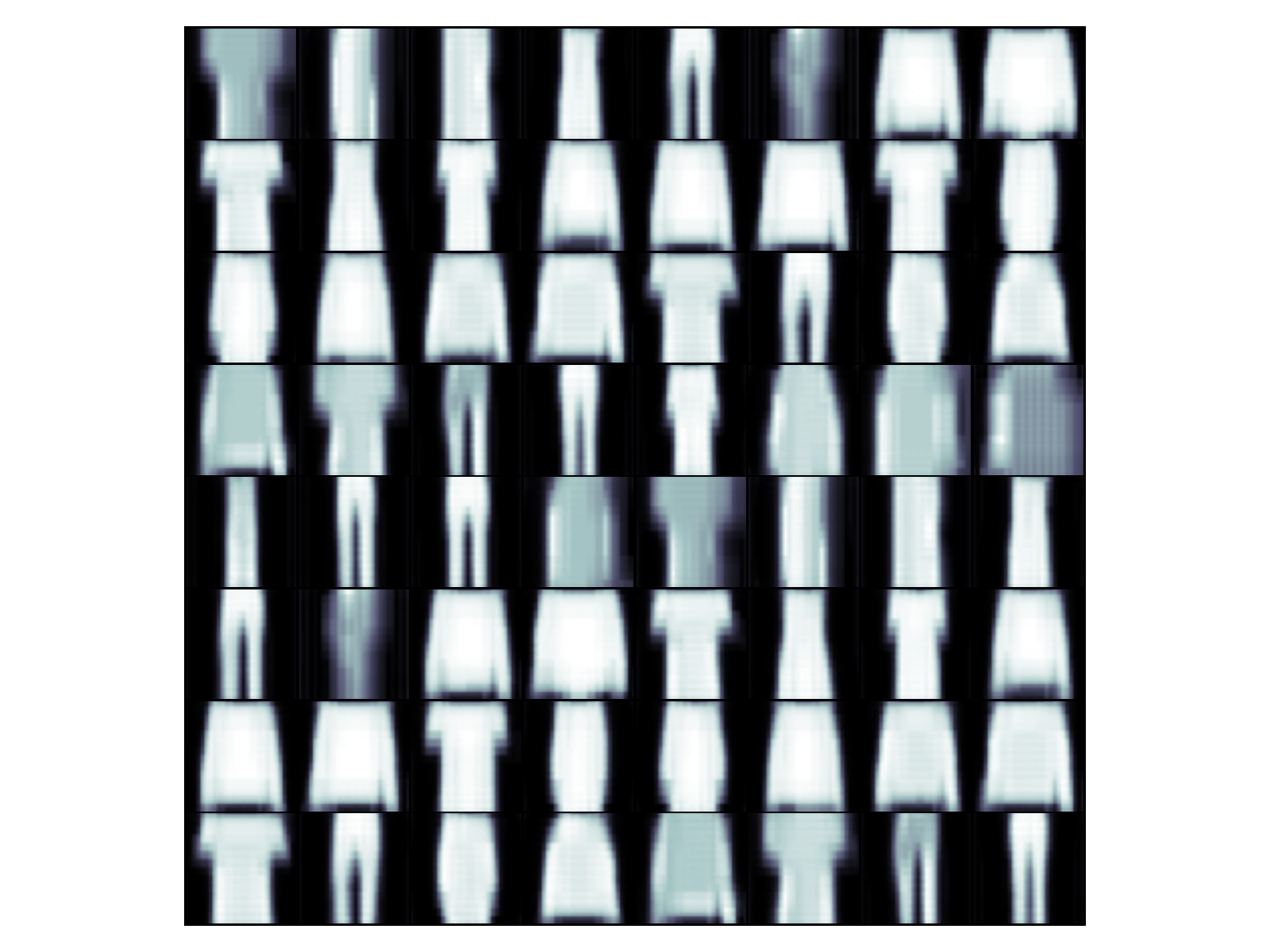}\subcaption{}\end{subfigure}\begin{subfigure}[c]{0.25\linewidth}\includegraphics[width=\linewidth,trim=2.1cm 0cm 2.1cm 0cm,clip]{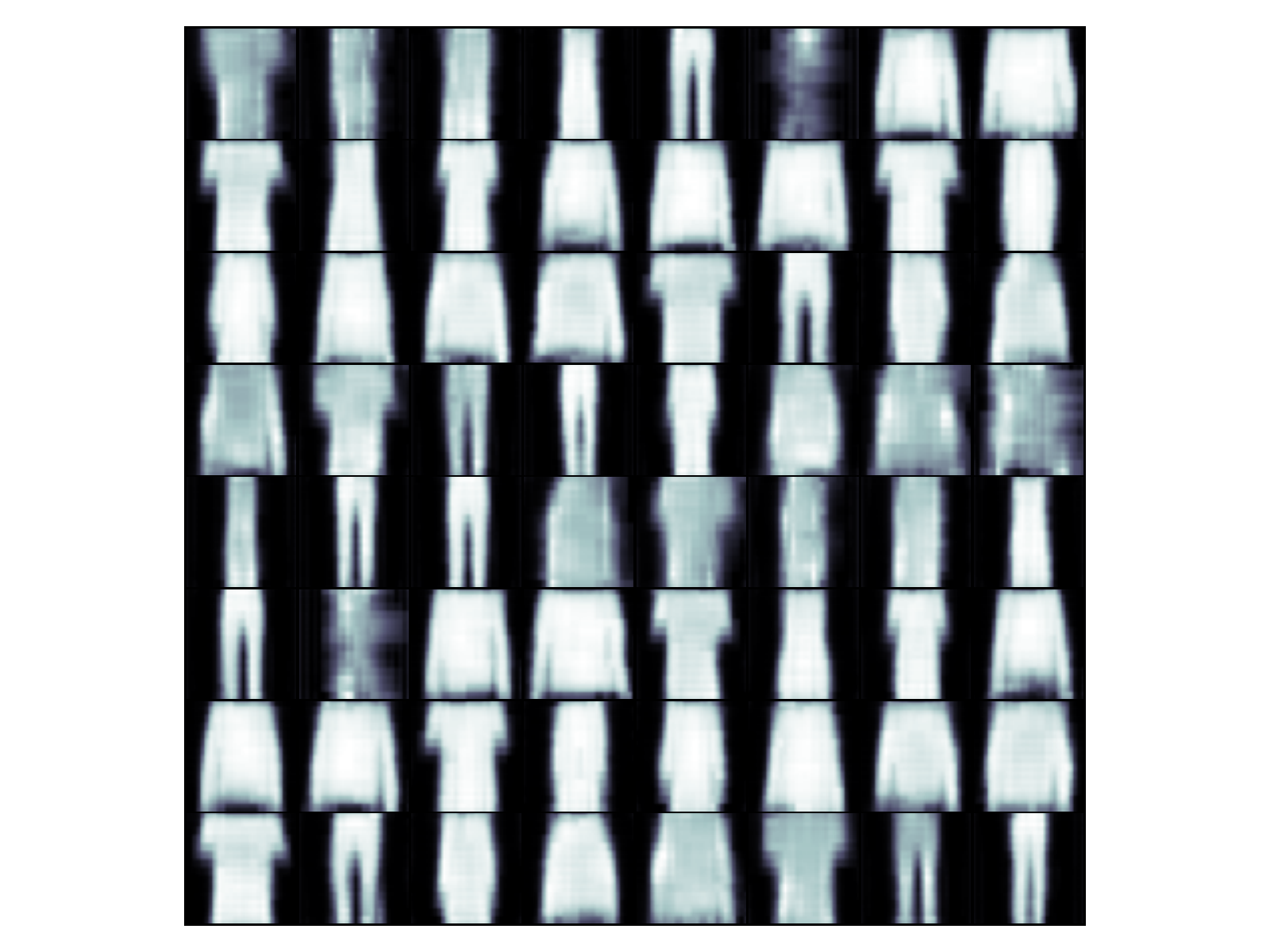}\subcaption{}\end{subfigure}
	\caption{
		Convolutional architecture is helpful for sampling: top-1 sampling shown on FashionMNIST, for \ac{DCGMM} architectures $1L$ (vanilla GMM, a), $2L$-$d$ (non-convolutional 2-layer, b), $2L$-$c$ and $2L$-$e$ (both convolutional 2-layer, c and d).
	}
	\label{app:samp5}
\end{figure}

\begin{figure}[htb]
	\centering
	\includegraphics[width=0.32\linewidth, trim= 2cm 0cm 2cm 0cm, clip]{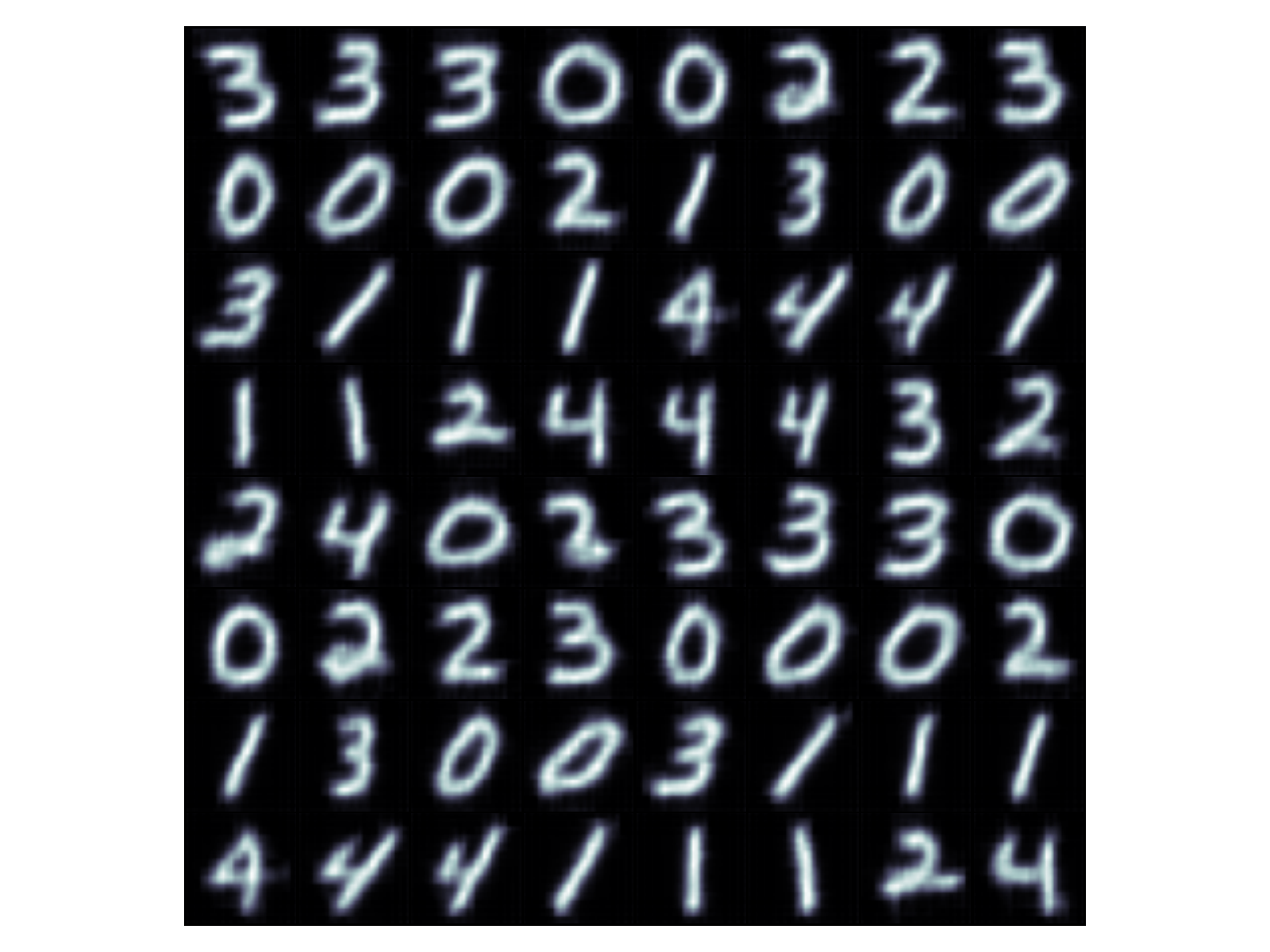}
	\includegraphics[width=0.32\linewidth, trim= 2cm 0cm 2cm 0cm, clip]{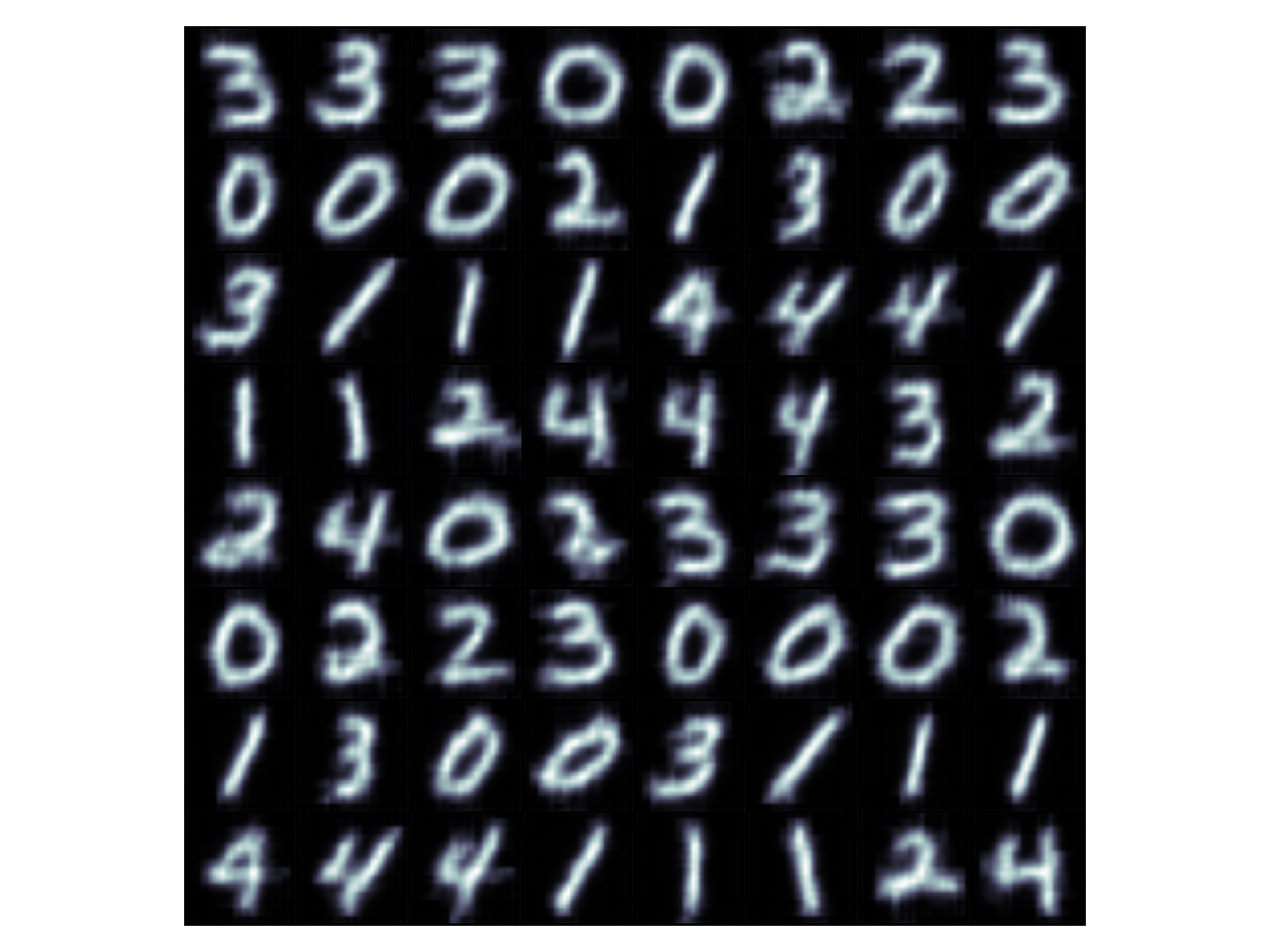}
	\includegraphics[width=0.32\linewidth, trim= 2cm 0cm 2cm 0cm, clip]{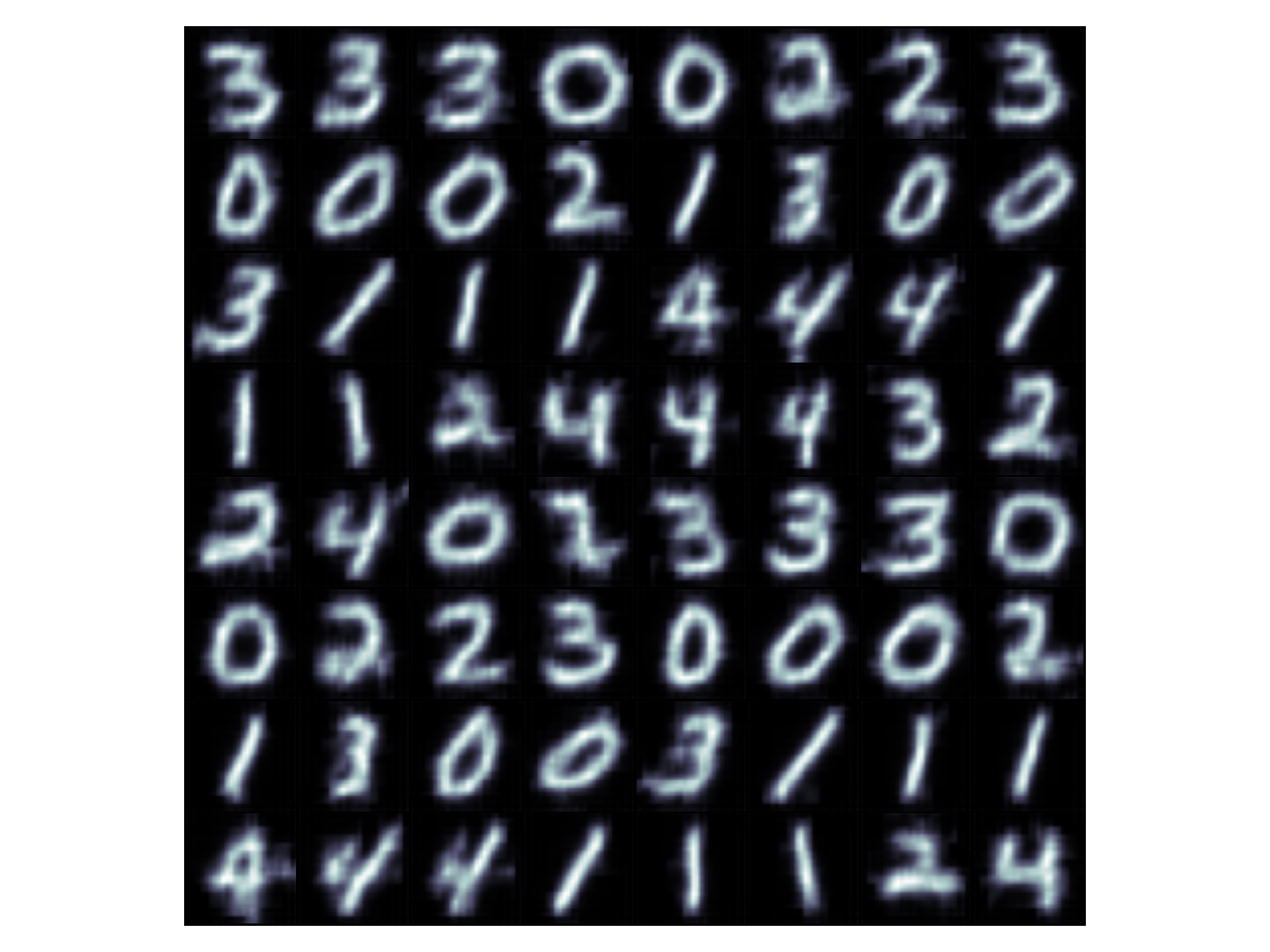}
	\includegraphics[width=0.32\linewidth, trim= 2cm 0cm 2cm 0cm, clip]{figs/mus2LcS2fm.pdf}
	\includegraphics[width=0.32\linewidth, trim= 2cm 0cm 2cm 0cm, clip]{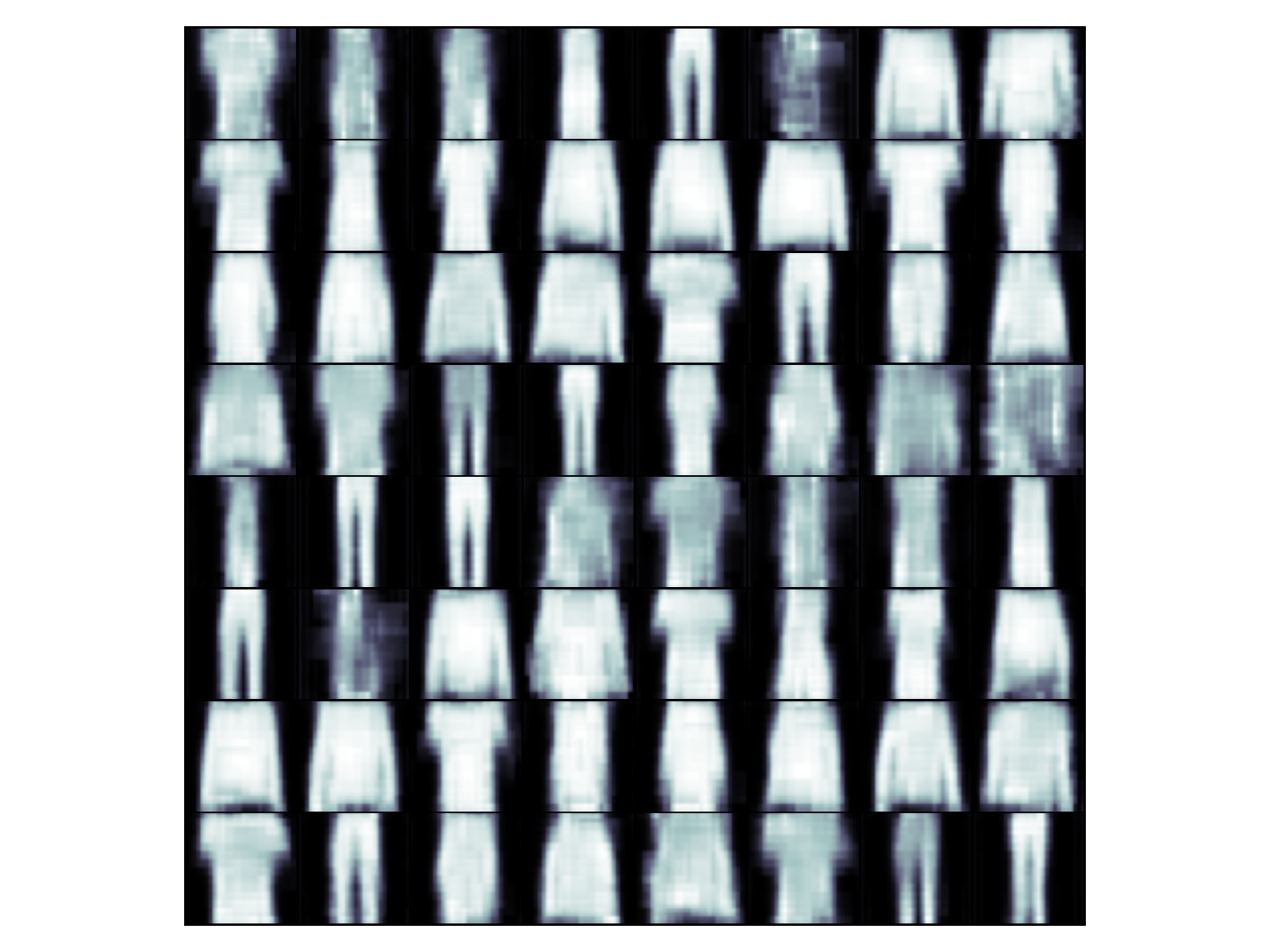}
	\includegraphics[width=0.32\linewidth, trim= 2cm 0cm 2cm 0cm, clip]{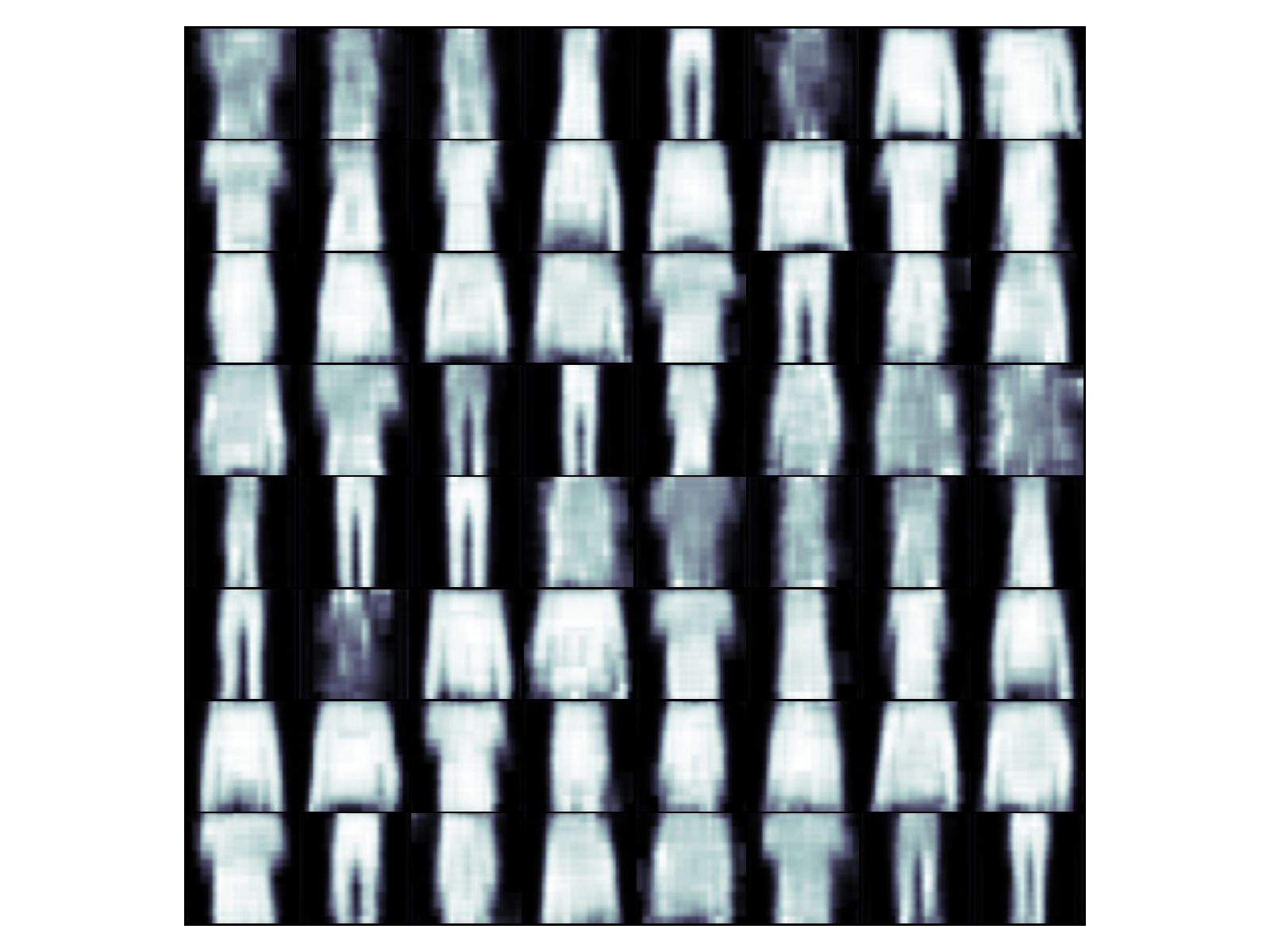}
	\caption{\label{fig:samp4}
		Impact of higher values of $S$ in top-S sampling, shown for \ac{DCGMM} instance $2L$-$c$ for MNIST and FashionMNIST. 
		From left to right: $S$=$2$,$5$,$10$.
	}
\end{figure}

\begin{figure}[htb!]
	\centering
	\includegraphics[width=0.32\linewidth,trim=2cm 0cm 2cm 0cm,clip]{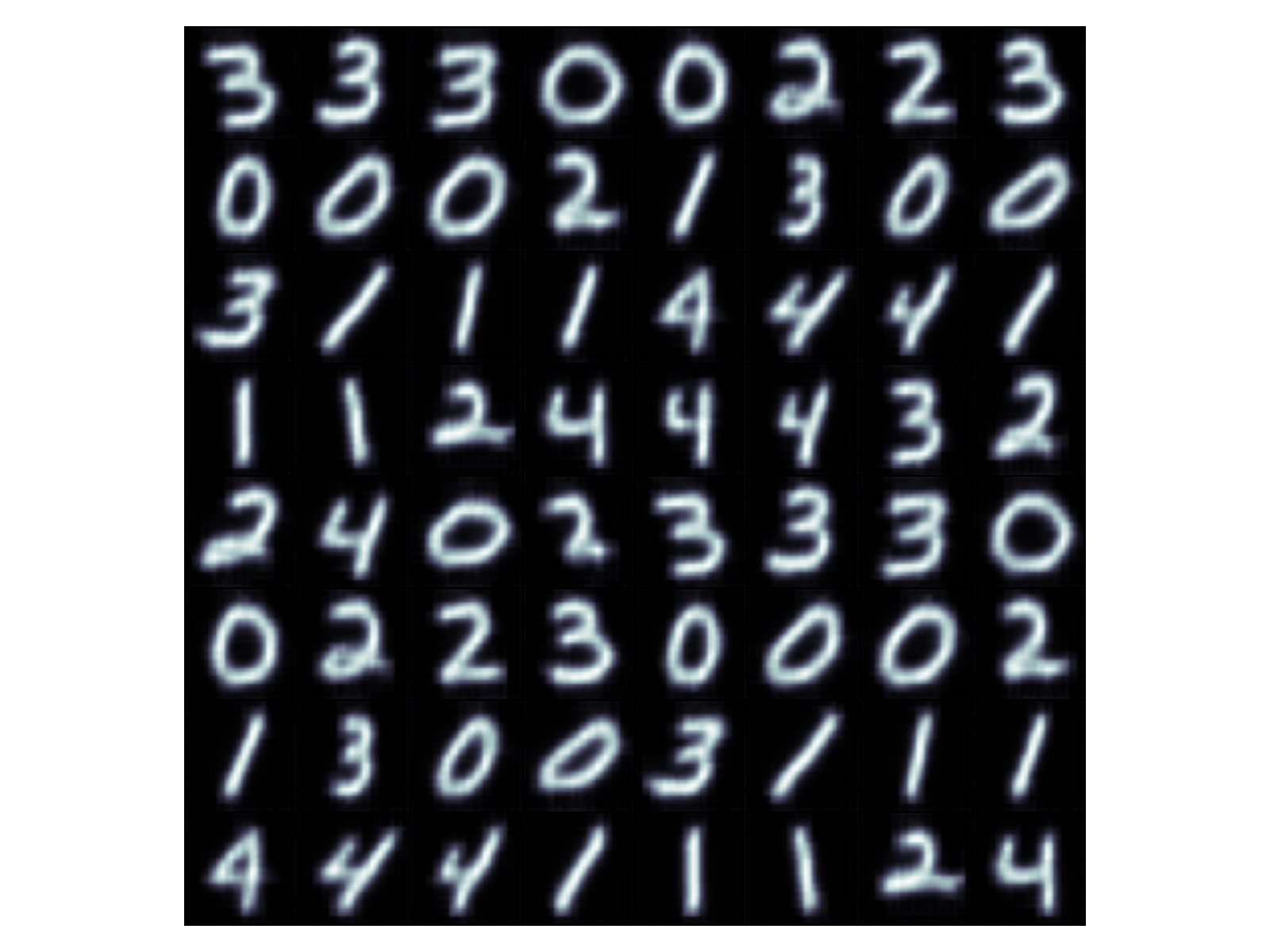}
	\includegraphics[width=0.32\linewidth,trim=2cm 0cm 2cm 0cm,clip]{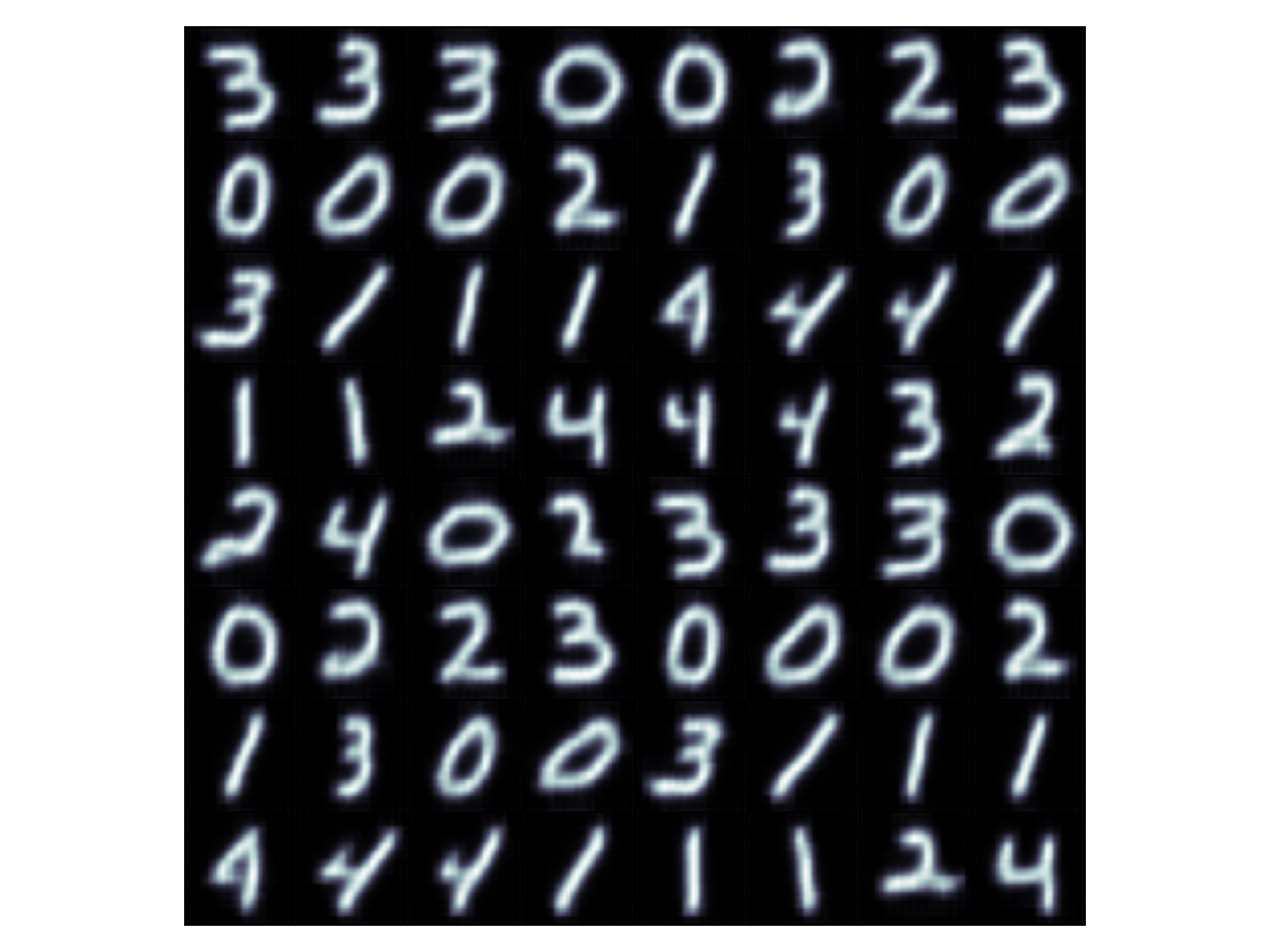}
	\includegraphics[width=0.32\linewidth,trim=2cm 0cm 2cm 0cm,clip]{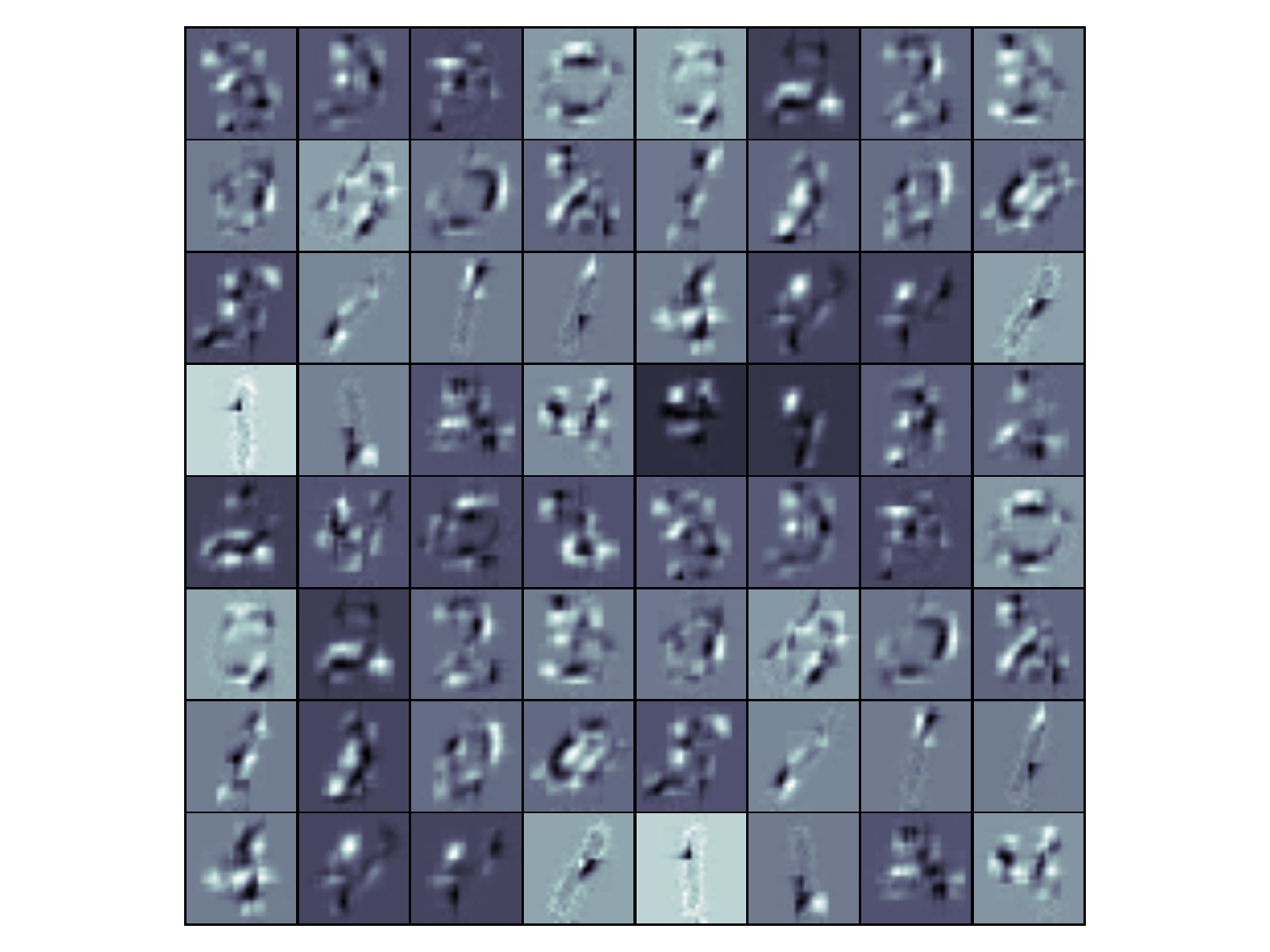}
	\includegraphics[width=0.32\linewidth,trim=2cm 0cm 2cm 0cm,clip]{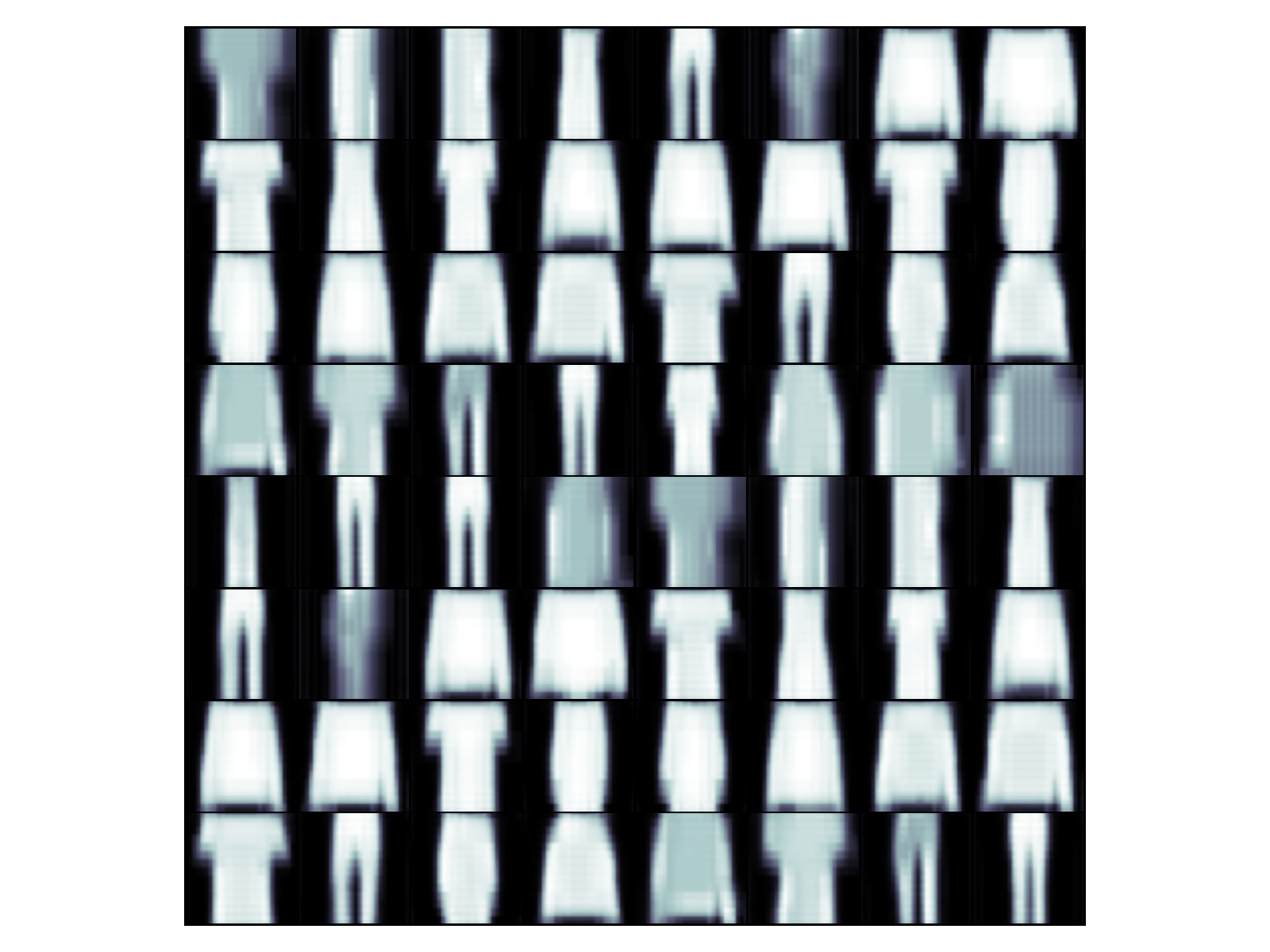}
	\includegraphics[width=0.32\linewidth,trim=2cm 0cm 2cm 0cm,clip]{figs/sharpening2Lcfm.pdf}
	\includegraphics[width=0.32\linewidth,trim=2cm 0cm 2cm 0cm,clip]{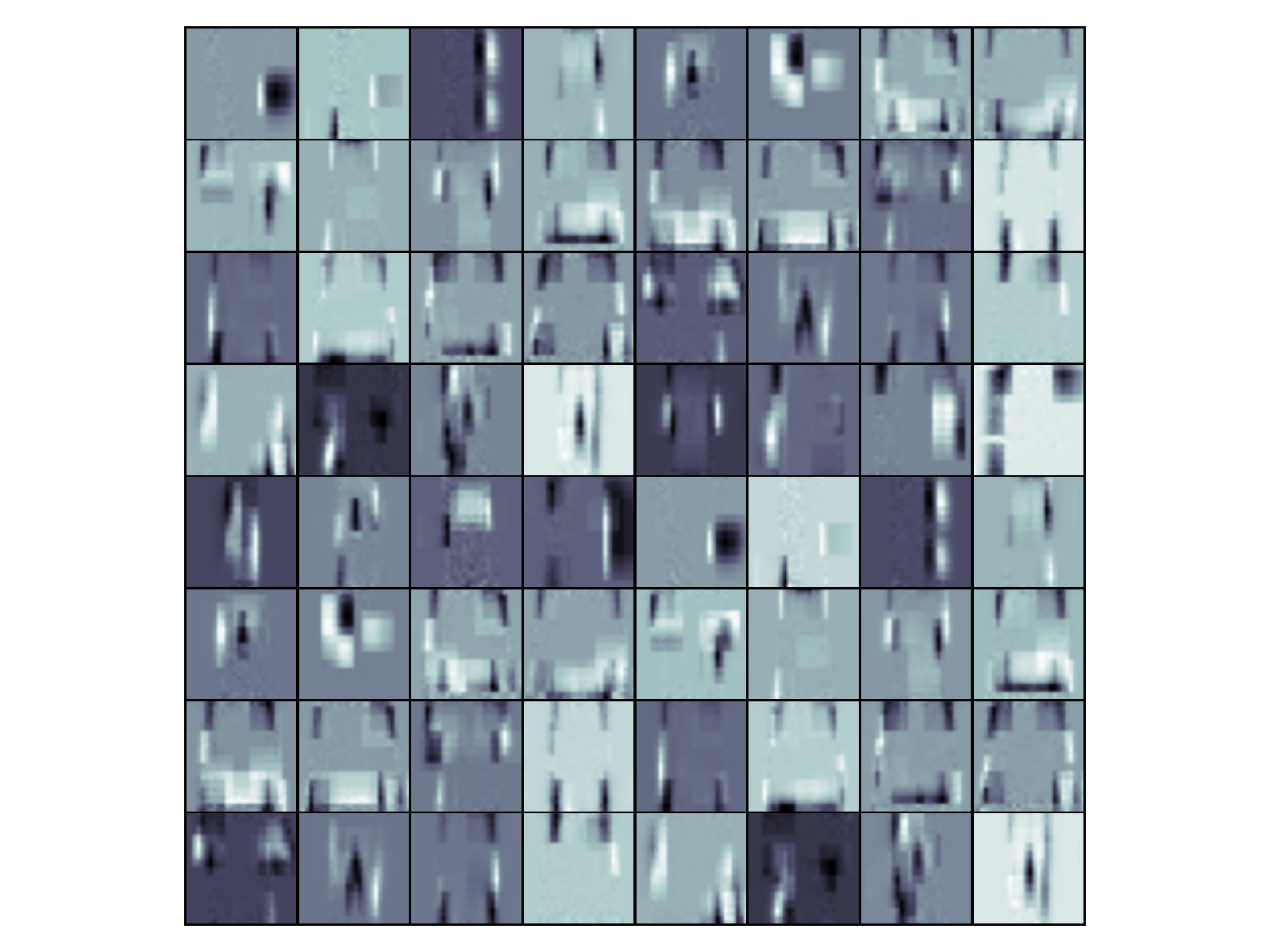}
	\caption{
		Impact of sharpening on top-S-sampling with $S$=$1$, see \Cref{sec:sampling}, shown for \ac{DCGMM} instance $2L$-$c$. 
		Shown are unsharpened samples (left), sharpened samples (middle) and differences (right). 
		Samples at the same position were generated by the same top-level prototype.
	}
	\label{fig:samp3}
\end{figure}

\paragraph{Effects of Convolutional Architectures}
An important property of \ac{DCGMM} is the ability to analyze \textit{local} input patches. 
In this experiment, we evaluate sampling for various two-layer architectures with different local patch sizes (i.e., convolution filter):
global ($2L$-$d$), large ($2L$-$a$), semi-local ($2L$-$c$) and local ($2L$-$e$) and observe effects when performing top-1-sampling.
The results of \cref{fig:global_local} indicate that as filter sizes decrease, the diversity but also the sharpness of generated samples increases at essentially no additional computational cost. 
Analogous FashionMNIST results are given in \cref{app:samp5}. 
\paragraph{Controlling Diversity by Top-S-Sampling}
Using instance $2L$-$c$, \cref{app:samp5} demonstrates how sample diversity is related to $S$. 
A higher value yields more diverse samples (see \cref{fig:samp4}), but increases the risk of generating corrupted samples or outliers. 
As the FashionMNIST results show, a good value of $S$ is clearly problem-dependent.
\paragraph{Generating Sharp Images}
\Cref{fig:samp3} shows the effect of sharpening for \ac{DCGMM} instance $2L$-$c$ using top-1-sampling. 
We can observe that the overall shape of a sample is not changed but that the outlines are sharper, an effect visible especially for FashionMNIST. 
Thus, sharpening does no harm and rather improves the visual quality of generated samples. 
\section{Probabilistic Interpretation of DCGMMs}\label{app:prob}
\noindent A probabilistic interpretation of the \ac{DCGMM} model is possible despite its complex structure.
The simple reason is that \ac{DCGMM} instances produce outputs which are inherently normalizable, meaning that the integral over an infinite domain (e.g., data space) remains finite.
Thus, \ac{DCGMM} outputs can be interpreted as a probability, which is not the case for \acs{DNN}/\acp{CNN} due to the use of scalar products.
\par
Here, we prove that GMMs are normalizable in the sense that the integral of the log-probability $\mathcal{L}(\vx)$\,$=$\,$\log \sum_k \pi_k p_k(\vx)$ is finite. 
This holds for any GMM layer in a hierarchy regardless of its input, provided that the input is finite (which is assured as Pooling and Folding layers cannot introduce infinities).
For simplicity, we integrate over the whole \mbox{$d$-dimensional} space $\R^d$. 
Since the component probabilities are Gaussian and strictly positive, and therefore the mixing weights are normalized and $\ge$\,$0$, the sum is strictly positive. 
Thus, it is sufficient to show that the integral over the inner sum (argument of the logarithm) is finite. 
We thus have
\begin{align}
	\begin{split}
		\int_{\R^d} \sum_k \pi_k p_k(\vx) d\vx & = \sum_k \pi_k \int_{\R^d} p_k(\vx)          \\
		                                       & = \sum_k \pi_k \sqrt{\text{det}(2\pi\Sigma)}
	\end{split}
\end{align} 
which is trivially finite because Gaussians are normalized.
\section{Conceptual and Practical Differences to Other Hierarchical GMM Models}\label{sec:diff}
\noindent In this work, we outline the main conceptual differences to what we consider the closest related work, namely the hierarchical GMM models presented in \cite{Viroli2019,VanDenOord2014,Tang2012}. 
Besides some differences, they share the notion that a GMM performs sampling by transforming what is termed a random \textit{latent variable}.
This process usually follows a simple distribution like $\mathcal N (0,\mI)$. 
\par
To clarify a purely semantic point: In this work, and close to the original derivation of the EM algorithm for GMMs \cite{Dempster1977}, we consider the latent variable $\vz$ of a GMM to be the one that enters into the complete-data log-likelihood of the corresponding latent-variable model. 
Another point where we differ in terminology is our perception of \enquote{lower} layers being closer to the input in density estimation mode. 
The lowest layer has index 1, counting upwards. 
\par
On the conceptual side, the cited models all share the notion that the proposed deep GMMs have a one-to-one correspondence to a flat model, which is actually the one whose log-likelihood is optimized. 
Consequently, layers are not optimized independently of each other, but are intrinsically linked. 
This is most easily visible from the fact that the component weights in each layer are not normalized.
Only the aggregated weights of all sampling paths through the deep model have this property.
\par
In contrast, our model does not assume (or require) a one-to-one correspondence to a flat GMM. 
The layers in our model are GMMs in their own right, having, e.g., their own, normalized weights and losses, and compute the posterior probabilities $\gamma_{nj} = p(z_j = 1 | \vx_n)$ of their \textit{own} latent variables. 
The dependency between layers is realized simply by the fact that posterior probabilities of one GMM layer are inputs to the subsequent GMM layer (potentially after being transformed by convolution and pooling layers). 
\par
The first practical consequence of this \enquote{independent layers} type of formulation is that layers can be optimized independently of each other. 
In particular, it is not necessary to enumerate all possible paths through the deep GMM as in \cite{VanDenOord2014} for training, which can only be approximated. 
Normal GMM optimization is rather performed for each layer, given the outputs of previous layers. 
This leads to a huge gain in scalability and allows deep GMMs with many layers to be trained on high-dimensional data, such as images, in a matter of minutes. 
\par
Another consequence concerns the introduction of pooling and convolution operations: since the layers of our model are not constrained by a joint \enquote{flat GMM} assumption, we are free to apply arbitrary transformations to their outputs before passing these on to subsequent GMM layers. 
The probabilistic interpretation of our model comes from the fact that each GMM in the model is independent and has the probabilistic interpretation that all GMMs share.
\par
A last consequence concerns sampling.
In our formulation of deep GMMs, a single GMM layer is trained on (estimated) latent variables from a preceding layer. 
When sampling, it can generate a valid instance $\vz$ of these latent variables. 
Therefore, it is reasonable to use the generated $\vz$ for \textit{selecting} the GMM component to sample from in layer $X$, instead of the $\vpi^{(X)}$. 
Of course, a suitable merging of both quantities is possible. 
Sampling, as described here, is a non-linear instead of an affine transformation, which is similar to the modeling related works.
Nevertheless, we retain the notion of \enquote{paths} through the DCGMM instance, given by the components selected for sampling in each layer.  
Due to the fact that the number of paths grows exponentially with the number of layers, deeper DCGMM instances can describe much richer sample distributions. 
\section{Summary, Discussion and Conclusion}
\noindent The \textbf{Objective of the article} was to establish the conceptual foundations of deep \ac{GMM} hierarchies (see also \Cref{app:prob}) that leverage important mechanisms from the \ac{CNN} domain. 
Conceptual differences and shared properties w.r.t.\ principal related works are discussed in \ref{sec:diff}.
Convolution and pooling layers make it possible to apply the model to high-dimensional image data with off-the-shelf hardware (typical training runs take approximately $2$ minutes on a GeForce GTX 1080). 
\par
\noindent\textbf{Results} show the illustration of important functionalities such as outlier detection, clustering and sampling, which no other work on hierarchical \acp{GMM} can present for such high-dimensional image datasets. 
We also propose a method to generate sharp images with GMMs, which has been a problem in the past \cite{Richardson2018}.
An interesting facet of our experimental results is that non-convolutional \acp{DCGMM} seem to perform better at clustering, whereas convolutional ones are better at outlier detection and sampling. 
\par
\noindent A \noindent\textbf{Key point} is the compositionality in natural images. 
This property is at the root of \ac{DCGMM}'s ability to produce realistic samples with relatively few parameters. 
When considering top-S-sampling in a layer $L$ with $H^{(L)}W^{(L)}$\,$=$\,$P^{(L)}$ positions, the number of distinct control signals generated layer $L$ is $S^{P^{(L)}}$. 
A \ac{DCGMM} instance with multiple \acl{G} layers $\{L_i\}$ can thus sample $\prod_L S^{P^{(L)}}$ different patterns, which grows with the depth of the hierarchy \textit{and} the number of distinct positions in a layer, making a strong argument in favor of deep convolutional hierarchies such as \ac{DCGMM}. 
This is an argument similar to the one about different \enquote{paths} through a hierarchical \ac{MFA} model in \cite{Viroli2019,VanDenOord2014}.
However, the number of paths grow more strongly for DCGMMs because sampling is performed independently for each GMM position.
\par
\noindent\textbf{Practical advantages} over other hierarchical models such as \cite{Viroli2019,VanDenOord2014,Tang2012} are most notably the introduction of convolution and pooling layers. 
Our experimental validation can be performed on high-dimensional data, such as images, with moderate computational cost.
This contradicts low-dimensional problems such as the artificial Smiley task or the \textit{Ecoli} and related problems.
Our experimental validation does not exclusively focus on clustering performance (problematic with images), but on demonstrating the capacity for realistic sampling and outlier detection.
Lastly, training \acp{DCGMM} by \ac{SGD} facilitates efficient parallelizable implementations, as demonstrated by the TF2 implementation provided.
\par
\noindent\textbf{Next steps} consist of exploring more \ac{DCGMM} architectures, mainly sampling, for generating natural images.
\clearpage
{\small
	\bibliographystyle{ieee_fullname}
	\bibliography{ijcnn2021}
}

\end{document}